%% file: main.tex
\documentclass[10pt,twocolumn,letterpaper]{article}

\usepackage{cvpr}              % To produce the CAMERA-READY version
\input{preamble}
\definecolor{cvprblue}{rgb}{0.21,0.49,0.74}
\usepackage[pagebackref,breaklinks,colorlinks,allcolors=cvprblue]{hyperref}

\title{Creative Image Generation with Diffusion Models}

\author{Kunpeng Song\\
Rutgers University\\
{\tt\small ks1418@scarletmail.rutgers.edu}
\and
Ahmed Elgammal\\
Rutgers University\\
{\tt\small elgammal@cs.rutgers.edu}
}

\begin{document}
\maketitle
\input{sec/0_abstract}    
\input{sec/1_intro}

\input{sec/2_Related_Work}
\input{sec/3_Motivation}
\input{sec/4_Preliminary}
\input{sec/5_Method}

\input{sec/6_Implementation}

\input{sec/7_Experiments}

\input{sec/8_Discussion}

\input{sec/9_Conclusion}

\clearpage

% --- Supplement numbering ---
\setcounter{section}{0}
\renewcommand{\thesection}{S\arabic{section}}
\renewcommand{\thesubsection}{S\arabic{section}.\arabic{subsection}}
\renewcommand{\thesubsubsection}{S\arabic{section}.\arabic{subsection}.\arabic{subsubsection}}

\input{sec/X_suppl}

\clearpage
{
    \small
    \bibliographystyle{ieeenat_fullname}
    \bibliography{main}
}
\end{document}

%% file: sec/0_abstract.tex
\begin{abstract}
Creative image generation has emerged as a compelling area of research, driven by the need to produce novel and high-quality images that expand the boundaries of imagination. In this work, we propose a novel framework for creative generation using diffusion models, where creativity is associated with the inverse probability of an image's existence in the CLIP embedding space. Unlike prior approaches that rely on a manual blending of concepts or exclusion of subcategories, our method calculates the probability distribution of generated images and drives it towards low-probability regions to produce rare, imaginative, and visually captivating outputs. We also introduce pullback mechanisms, achieving high creativity without sacrificing visual fidelity. Extensive experiments on text-to-image diffusion models demonstrate the effectiveness and efficiency of our creative generation framework, showcasing its ability to produce unique, novel, and thought-provoking images. This work provides a new perspective on creativity in generative models, offering a principled method to foster innovation in visual content synthesis. 
\footnote{Project page: https://creative-t2i.github.io/}
\end{abstract}

%% file: sec/1_intro.tex
\section{Introduction}
\label{sec:intro}

\begin{figure*}[ht]
\vskip 0.2in
\begin{center}
\includegraphics[width=1\linewidth]{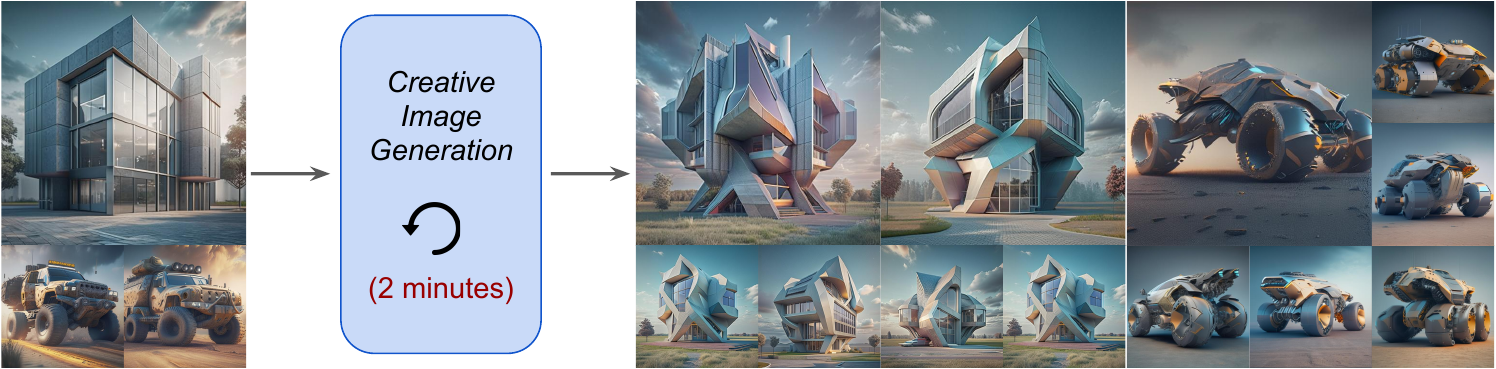}
\caption{Creative Generation from our method for building and vehicle, taking only 2 minutes.}
\label{hero_c}
\end{center}
\vskip -0.2in   
\end{figure*}

With the emergence of text-to-image generative models like Stable Diffusion \cite{rombach2022high,esser2024scaling}, SDXL \cite{podell2023sdxl}, and DALL \cite{ramesh2021zero,ramesh2022hierarchical}, we now have unprecedented tools to transform language into diverse, high-quality images. These advancements have enabled a wide range of applications, including text-to-image generation, image editing, etc. Recent personalization techniques have further enhanced the ability to customize and adapt generative processes. 

% Human creativity remains a hallmark of our species, but its definition often varies across disciplines. Despite the historical and ongoing debate about what creativity is, it is always characterized in terms of two qualities: On one side, there is novelty and surprise, and on the other side is value, or influence~/cite{paul2014philosophy}. While novelty is rooted in the uniqueness perceived by the creator or the observer, the concept of "value" is more nebulous. Is it the market value, aesthetic appeal, or the influence of the artwork? Various definitions have tried to encapsulate this concept.

Creativity remains one of the hallmarks of our species. It's associated with being rare and surprising and is rooted in the uniqueness perceived by the creator or the observer. Now, a critical question is: \textit{\textcolor{BlueViolet}{Can these powerful AI generate truly novel and creative outcomes?}} In other words, \textit{\textcolor{BlueViolet}{Are generative AI models capable of being truly creative?}}

% The critical questions are: how creative are these cutting-edge generative AI systems, and how do users interact creatively with them? What is the source of creativity when a user collaborates with such systems? We argue that these systems fundamentally lack creativity for several reasons. By construction, they are designed to sample from input distributions, with their inherent loss functions displaying a pronounced bias towards typical outputs. When it comes to evaluation, the typical metrics employed tend to prioritize how closely generated images aligns with the training data, inadvertently fostering a counter-creative bias, favoring replication over innovation. In terms of usability, these AI models heavily depend on human input as the source of creativity. Users often resort to a trial-and-error method and prompt engineering, leaning on their own creativity to derive meaningful or unique outcomes from the system.

We argue that these systems fundamentally lack creativity for good reasons. They are designed to mimic training distributions, with their loss functions displaying a pronounced bias toward typical outputs. When it comes to evaluation, the typical metrics tend to prioritize how closely generated images aligns with the training data, inadvertently fostering a counter-creative bias, favoring replication over innovation. Basically, image generating models are mainly focused on improving image quality. This is an instance of the quality/creativity trade off, where it prioritizes image quality by following the training data distribution. These AI models are heavily biased toward typical outputs, which are intrinsically boring. Users often depend on human input as the source of creativity and resort to prompt engineering, leaning on human creativity to derive novel or unique outcomes from the system.

In this paper, We present a generative AI system that is encouraged to generate novel images. We address the task of creative text-to-image generation, where creativity is defined as the ability to produce novel, high-quality outputs that exhibit a low probability of existence. Unlike methods that explicitly blend or combine concepts \cite{richardson2023conceptlab,elgammal2017can, xiong2024novel}, our method samples from low-probability regions in the generative model's output space, fostering creativity without manual intervention. For example, within the category of “handbag,” we aim to generate images that semantically resemble a handbag while differing from any known common norms. This probabilistic framework provides an effective approach to creativity, pushing the boundaries of generative models and unlocking new possibilities for imaginative visual content. Specifically, our contributions are as follows:

% \begin{itemize}
% \item New perspective on creative AI. We view creativity in the context of text-to-image generation tasks, by associating it with probability.
% \item Creativity-oriented optimization. We design a specialized deep learning system with a creativity-driven loss function to directly optimize for low-probability outputs, enabling the generation of visually captivating and novel content.
% \item Creative Image Generation Pipeline. We build a creative image generation pipeline, leveraging two state-of-the-art diffusion models, Kandinsky 2.1 \cite{kandinsky2}. 
% \item Directional Control for Enhanced Guidance. We develop a directional control method to add more precise and flexible controls to the generative process, allowing for better alignment between user-defined objectives and the generated outputs.
% \end{itemize}
\begin{itemize}
    \item \textbf{New perspective on creative AI.} We view creative image generation through the lens of probability, explicitly targeting low-probability regions to foster novel outputs.
    \item \textbf{Creativity-oriented optimization.} We design a specialized loss function that directly encourages the exploration of less probable image embeddings, driving the model toward more creative and imaginative results.
    % \item \textbf{Creative image generation pipeline.} We build upon the diffusion-based framework of Kandinsky~2.1, integrating our creativity-driven approach to produce a flexible, high-fidelity generation pipeline.
    \item \textbf{Pullback and Directional control.} We introduce pullback constraints that guard against out-of-domain collapse. And a method to steer the model’s exploration trajectory in specific directions, while maintaining both creativity and semantic fidelity.
\end{itemize}

%% file: sec/2_Related_Work.tex
\section{Related Work}
\subsection{Counter-creative Bias in Generative Systems}

% reduced version
Image-generating models aim to produce images resembling training data, often lacking mechanisms to explore creative spaces. Traditional GANs sample from the data distribution~\cite{Goodfellow2014} but struggle with balancing quality and diversity, leading to issues like mode collapse and mode dropping. To address this, techniques such as truncation in BigGAN~\cite{Brock2018} adjust latent vectors to prioritize quality over diversity. Other generative models, including VAEs~\cite{Kingma2013}, flow-based methods~\cite{dinh2014nice,kingma2018glow}, and score-matching models~\cite{song2020score}, also primarily replicate training distributions rather than fostering creativity. VAEs sample from learned latent spaces, flow-based models transform base distributions, and score-matching estimates data gradients. Diffusion models refine image generation through denoising~\cite{ho2020denoising}, with guided approaches enhancing quality, often via text prompts~\cite{ramesh2022hierarchical,ho2022classifier,rombach2022high} . However, like other methods, they optimize for image reconstruction without incentives to explore beyond the training data.

\subsection{Counter-creativity Bias in Evaluation Metrics}
Standard metrics for evaluating Generative AI systems prioritize image quality over creativity. The Fréchet Inception Distance (FID) is a popular metric in this context~\cite{heusel2017gans,theis2015note}. It gauges the similarity between real and generated image distributions. 
% A lower FID suggests better quality, while a higher one indicates divergence. 
% Therefore, 
FID is a ``typicality'' metric, favoring conventional images. Systems with superior FID scores are less likely to produce innovative results. 
Another metric, the Inception Score (IS), evaluates image generative systems based on quality and diversity~\cite{salimans2016improved}. 
% A higher IS indicates both quality and diversity. However, while 
while it ensures representation across sub-categories, it doesn't promote novelty within or beyond these subcategories.

\subsection{Related work on Creative Generative systems}
Before the rise of modern generative AI, computational creativity literature proposed algorithms to effectively navigate the creative space. Many employed evolutionary processes, generating candidates, assessing them via a fitness function, and refining them for subsequent iterations, often within a genetic algorithm framework~\cite{machado2008iterative,dipaola2007incorporating}. The challenge was crafting a logical fitness function with aesthetic sensibilities. Some systems incorporated human feedback, with humans guiding the creative exploration~\cite{baker1994evolving,graf1995interactive}. Recent systems have underscored the importance of perception and cognition in creativity~\cite{colton2008creativity}. 

In the GAN context, CAN~\cite{elgammal2017can} demonstrated modifying GAN loss to encourage creative exploration. This involved creating tension between adhering to general art distribution and producing unique and novel art styles. This tension nudged the system towards novel creations, addressing the novelty/value balance.  In the realm of Text-to-image diffusion, ConceptLab~\cite{richardson2023conceptlab} was introduced recently to push the system for novel concepts and styles. By setting prompt constraints, like generating a pet image that isn't a cat, dog, or hamster, the system is indirectly pushed towards creativity. Both CAN and ConceptLab exploits the existence of subcategories within the general concept to push the system to generate novel images that belong to the concept but not to the subcategory. In contrast, our proposed approach does not assume or rely on existence of such subcategories and directly aim at optimizing a creative loss. 

%take a look:
% Prior work ConceptLab generates creative outputs by inducing classification uncertainty, encouraging models to deviate from well-defined subcategories or style norms. However, this mechanism inherently relies on having clear subcategories or labels to serve as a reference for measuring novelty. Without such structured benchmarks, the feedback loop that guides the model toward truly creative (yet coherent) outputs is lost, making it challenging to evaluate or steer the generated content away from common modes.
%In summary, while strides have been made, there's a significant gap in integrating creative losses into GenAI systems, a deficiency we aim to address in this

\subsection{Diffusion-Based Generative Models}
Large-scale text-to-image diffusion models \cite{dhariwal2021diffusion, ho2020denoising, nichol2021improved} have achieved an unprecedented ability to generate high-quality imagery guided by a text prompt \cite{nichol2021glide, ramesh2022hierarchical, rombach2022high, saharia2022photorealistic, kandinsky2}. Leveraging these powerful generative models, many have attempted to utilize such models for downstream editing tasks \cite{couairon2022diffedit, hertz2022prompt, meng2021sdedit}. Most text-guided generation techniques condition the model directly on embeddings extracting from a pretrained text encoder \cite{rombach2022high, kandinsky3}. In this work, we utilize Kandinsky 2.1 \cite{kandinsky2} a Latent Diffusion Model that consists of a diffusion prior and a diffusion UNet. 

%% file: sec/3_Motivation.tex
\section{Motivation and Justification}
\label{motivation}

\noindent{\bf Background:}
In Psychology literature, D. E. Berlyne (1924-1976) emphasized the role of “arousal” in aesthetics, defining it as a measure of alertness, ranging from relaxation to intense excitement~\cite{Berlyne1967,Berlyne1971}. Arousal potential refers to stimulus properties that increase arousal, with novelty, surprisingness, complexity, ambiguity, and puzzlingness being the most significant for aesthetics. He termed these collative variables.  Studies show a preference for moderate arousal potential, as too little is boring, while excessive arousal activates aversion, reducing hedonic response~\cite{Berlyne1967,schneirla1959evolutionary}. This relation is captured by the Wundt curve~\cite{Berlyne1971,wundt1874grundzuge}, though alternative models have been proposed~\cite{kuppens2013relation}.  

\noindent{\bf Novelty and Surprisingness:}
\label{arousal_probability}
Novelty in creative generation differs from merely producing unseen samples. In high-dimensional spaces used in GANs and diffusion models, samples near the mean are novel but often typical. Creativity requires deviation from the mean to enhance arousal potential. A creative system must balance novelty and value~\cite{Boden2010}. Increasing novelty involves sampling from low-probability regions, but excessive deviation lowers perceived quality. Thus, a "Pullback" mechanisms are needed to regulate novelty without compromising value.  

\noindent{\bf Increasing Arousal Potential of Samples:}
Berlyne’s model ties novelty to deviation from prior experiences. We focus on novelty as a driver of arousal potential to enhance creativity.  

In a simulated setting, novelty can be quantified using information theory, considering a user's prior exposure. Since direct measurement is impractical, we approximate exposure by sampling images related to a user prompt, forming a model $M$. Arousal Potential (AP) is then estimated as:  
\[
AP_{novelty} (x | M) = - \log(P(x | M))
\]
From an information theory perspective, novelty aligns with surprise. While distinct, for this paper, we assume equivalence since statistical expectation can define surprise based on training data.

% Creativity Metrics and Novelty-Driven Generation
% The Creative Adversarial Network (CAN) \cite{elgammal2017can} generates creative art images by training the generator to produce images that the discriminator accepts as "art" but cannot classify into any specific style. This forces the generator to deviate from established styles, fostering creativity through novelty and style deviation. Weaver \cite{wang2024weaver} details pre-training on a carefully selected corpus to train domain-specific LLMs for creative and professional writing purposes.

% \subsection{Multimodal Vision-Language Models}

%% file: sec/4_Preliminary.tex
\section{Preliminary}
\label{preliminaries}
\noindent\textbf{Latent Diffusion models} 
Latent Diffusion Models (LDMs) aim to generate images by performing the diffusion process in a compressed latent space. Formally, let 
\(
x \in \mathbb{R}^{H \times W \times 3}
\) 
be an image, and let 
\(\mathcal{E}\) 
and 
\(\mathcal{D}\) 
denote the encoder and decoder of an autoencoder, respectively. The encoder 
\(\mathcal{E}\) 
maps 
\(x\) 
to a latent representation 
\(
z = \mathcal{E}(x)
\) 
, while the decoder 
\(\mathcal{D}\)  
reconstructs 
it such that 
\(\mathcal{D}(z) \approx x\).

A diffusion model is then trained on the latent codes \(\{z\}\) using DDPM \cite{ho2020denoising} objective. Let 
\(z_t\) 
be the latent code at diffusion timestep 
\(t\).
The network 
\(\epsilon_\theta\) 
learns to predict the injected noise 
\(\epsilon\) 
at each step, conditioned on a latent code 
\(
z_t, 
\) 
the timestep 
\(t,\) 
and an optional conditioning vector 
\(c\):
\begin{equation}
\label{eq:ldm}
\mathcal{L}_\text{LDM} \;=\; \mathbb{E}_{z,\epsilon,t}\bigl[\|\epsilon \;-\; \epsilon_\theta(z_t,\,t,\,c)\|^2 \bigr],
\end{equation}
% When training, \(\epsilon_\theta\) becomes proficient at progressive denoising \(z_t\) into a clean latent code \(z_0\). At inference, LDMs sample in the latent space and decode it to an image via \(\mathcal{D}\).

% By shifting diffusion to a more compact latent space, LDMs can reduce computational demands while preserving fidelity. Moreover, because the autoencoder and diffusion components are trained separately, one can fine-tune or augment specific parts (e.g., the diffusion process) without retraining the entire pixel-level model.

\noindent\textbf{Text-to-Image diffusion prior}
% Kandinsky diffusion prior
Our method is built on \emph{Kandinsky 2.1} \cite{kandinsky2} which decomposes text-to-image generation into two stages:
(i)~a diffusion prior \(\epsilon_\theta\) that predicts CLIP image embeddings $e \in \mathbb{R}^m$ from text prompts, 
(ii)~a diffusion decoder that takes in image embeddings to generate images. Formally, the diffusion prior \(\epsilon_\theta\) is trained on:
\begin{equation}
\label{eq:prior}
\mathcal{L}_\text{prior} \;=\; \mathbb{E}_{e,\epsilon,t}\bigl[\|\epsilon \;-\; \epsilon_\theta(e_t,\,t,\,\phi(P))\|^2 \bigr],
\end{equation}
,where \(P\) is a text prompt, \(\phi\) the text encoder, \(e_t\) the noised image embeddings. This two-stage approach allows flexible control over the intermediate CLIP image embedding. 

% Our method works largely in this CLIP space.
% , driving it toward creative outputs.

% %\(\phi(y)\in \mathbb{R}^d\). 
% % instead of directly generating an image from 
% % \(\phi(y)\), 
% , and  a diffusion prior. An image embedding 
% \(
% e \in \mathbb{R}^m
% \) 
% aligned to the latent representation space (e.g., a CLIP embedding space). Concretely, at each timestep 
% \(t\), 
% the prior models:
% \[
% e_{t-1} \;\leftarrow\; P_\theta(e_t,\,t,\,\phi(y)),
% \]
% where noise is gradually removed from an initial random embedding. 
% Once trained, the prior can stochastically generate plausible image embeddings 
% \(\hat{e}\) 
% that is subsequently fed into a separate diffusion decoder (such as the one in Section~\ref{eq:ldm}) to render a final RGB image.

% In our work, we employ the \emph{Kandinsky 2.1} prior, which Kandinsky 2.1 

% lower-probability yet visually coherent, enabling the emergence of more creative outputs. 

%% file: sec/5_Method.tex
\section{Method}

\begin{figure*}[ht]
\vskip 0.2in
\begin{center}
\includegraphics[width=1.0\linewidth]{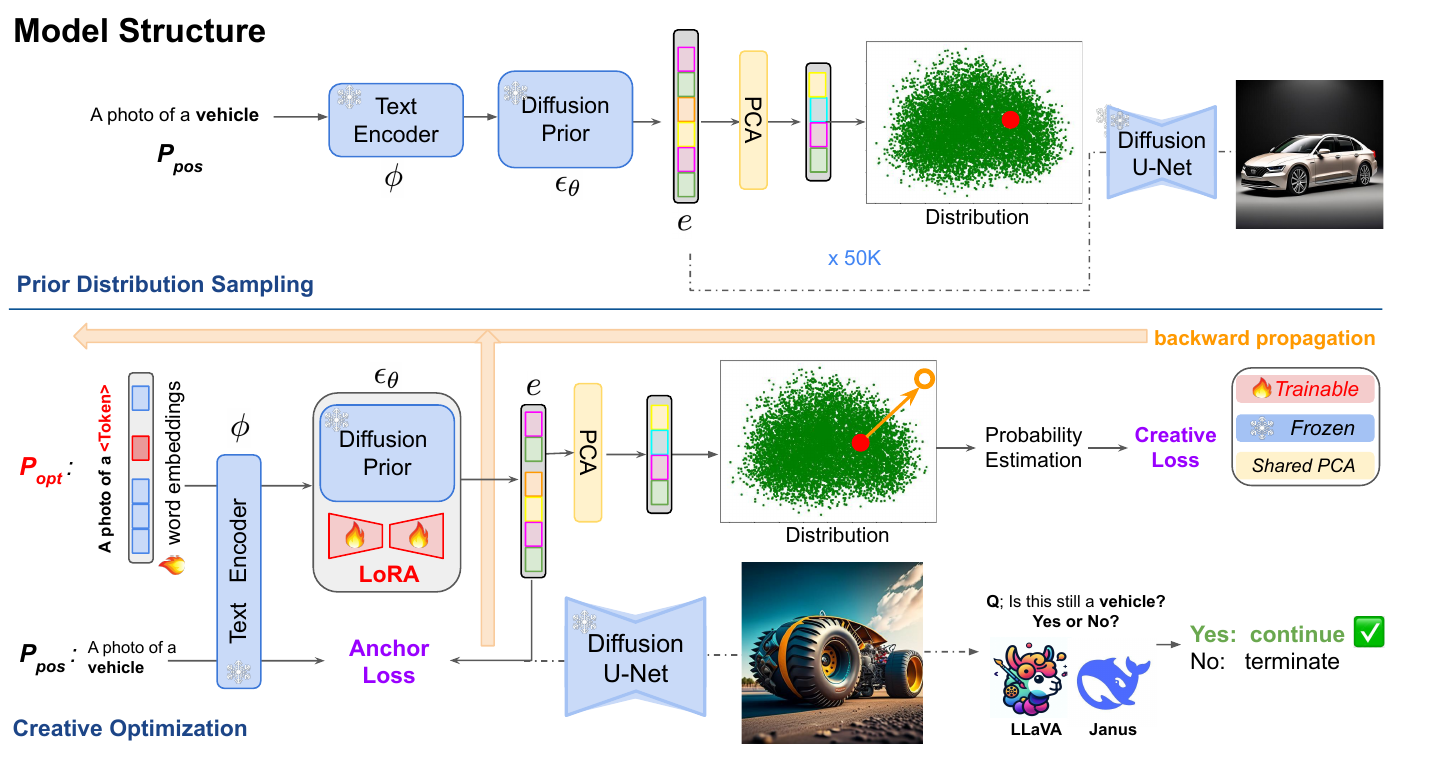}
\caption{Overall Model Structure. We first sample a distribution \textcolor{OliveGreen}{(green cluster)} of the generated image embeddings \(e\) from the diffusion prior \(\epsilon_\theta\) (top). Then, during creative optimization (bottom), we optimize learned token and LoRA layers with creative loss to push the generated embeddings \textcolor{red}{(red dot)} toward low-probability regions \textcolor{orange}{(orange arrow)}, constrained by an anchor loss and validated by a multimodal LLM. The diffusion decoder finally renders the resulting images. 
% Dash lines mean no gradient is required.
}
\label{icml-historical_2}
\end{center}
\vskip -0.2in
\end{figure*}

We now detail our model design for creative text-to-image generation. We can identify four components of a creative generative system: 1) Conceptual space, 2) Optimization criteria, 3) Pullback mechanism, 4) Directionality. At a high level, see Figure~\ref{icml-historical_2}, we first learn a \textit{baseline} distribution of image embeddings \(e\) from the diffusion prior \(\epsilon_\theta\), as detailed in \ref{PPS}. Next, we encourage exploration of \textit{low-probability} regions in that distribution, as detailed in \ref{CO}. These combined can increase the likelihood of producing novel or ``creative'' outputs. Then in \ref{PB}, we ensure semantic validity
% (e.g., ``still looks like a building'') 
by applying pull-back constraints including a positive anchor and an MLLM semantic validity checker, which effectively prevent out-of-domain failures. Finally, we introduce directionality controls via negative clusters, as detailed in \ref{NC}. Now we elaborate on each components:
% querying a multimodal large language model

\subsection{Prior Distribution Sampling}
\label{PPS}
As mentioned in section \ref{preliminaries},   $e \in \mathbb{R}^m$ denotes the image embedding generated by diffusion prior \(\epsilon_\theta\), given a text prompt $P$. We draw a large number of samples $\{e_i\}_{i=1}^{N}$ 
% (e.g., $N=5{,}000$) 
to approximate the \textit{baseline} distribution of generated image embeddings induced by the prior \(\epsilon_\theta\). To reduce dimensionality, we apply principal component analysis (PCA)~\cite{doi:10.1080/14786440109462720}, producing 
% lower-dimensional embeddings 
$\tilde{e} \in \mathbb{R}^k$ ($k \ll m$). We then fit a multivariate Gaussian $\hat{G}(\tilde{e})$ to these PCA-transformed samples. Formally:

\begin{equation}
    \tilde{e} = \mathbf{W}(e - \mathbf{\mu_0}); 
\end{equation}
\begin{equation}
    \hat{G}(\tilde{e}) = \mathcal{N}\bigl(\tilde{e}\,\big|\;\mathbf{0},\,\mathbf{\Sigma}\bigr),
\end{equation}
,where $\mathbf{W}$ is the PCA projection matrix, $\mathbf{\mu_0}$ is the mean of the original embeddings, and $\mathbf{\Sigma}$ is the estimated covariance in the reduced space.
% For simplicity, we write $\tilde{e} = pca(e)$.

\subsection{Creative Optimization}
\label{CO}
\noindent\textbf{Conceptual space}
Conceptual space refers to the parameter space that a creative system is  searching in during optimization~\cite{Boden2004}. 
% Unlike previous works, which explored Latent space (CAN~\cite{elgammal2017can}) or Text embedding space (ConceptLab \cite{richardson2023conceptlab}),
% (as in GAN % , VAE, flow-based models).
% 2) Image embedding space (like CLIP [Radford 2021] and BLIP [Li 22022]). 
We use a combined space of token embedding and Low-Rank Adaptation \textbf{(LoRA)}~\cite{hu2022lora} parameters as the conceptual space for creative exploration. Specifically, We optimize both a token embedding, eg. vehicle, and LoRA rank-decomposition matrices of the diffusion prior \(\epsilon_\theta\). We use \(P_{pos}\) as the positive prompt\footnote{An example of prompts \(P_{pos}\) we use: Professional high-quality photo of a vehicle. photorealistic, 4k, HQ.}, e.g., "a photo of a vehicle". And \(P_{opt}\) as the prompt containing the token to be optimized, 
e.g., "a photo of a $<token>$". This learnt token embedding is initialized using the default subject token, e.g., "vehicle". For LoRA parameters, we randomly initialize all matrix A, and zero initialize all matrix B. 
% for efficient and effective optimizations.

\noindent\textbf{Creative Loss}
We approach creative image generation by increasing its arousal potential (see Sec.\ref{arousal_probability}).
% How the text embedding is mapped to a sample and compared to distribution to compute the sample arousal potential and use that as the . 
% Such loss is then optimized and provided the directions to change sample in the conceptual space,  the text token embedding. 
Our key insight is that \textit{being creative} requires moving \textit{away} from high-probability embedding regions. To do so, we define a \textbf{creative loss} function that encourages the dimension-reduced predicted embedding $\tilde{e}$ to reside in regions of lower probability under $\hat{G}$. Specifically, minimizing its log-likelihood,

\begin{equation}
    \mathcal{L}_{\text{creative}}(\tilde{e}) = \log \hat{G}(\tilde{e}).
\end{equation}

\noindent$\mathcal{L}_{\text{creative}}$ forces
$\tilde{e}$ toward the tails of the distribution $\hat{G}$, increasing the chance of getting novel, creative samples.

% $\log \hat{G}(\tilde{e})$ to \textit{decrease} and thus drives 

% \textcolor{red}{emphasize the difference between our model and conceptLab, that ours works without having to rely on subclasses.}

% We backpropagate the gradient through the PCA projection and diffusion prior’s LoRA parameters into text embeddings, iteratively updating them such that the model produces lower probability embeddings.

% \subsection{Pullback mechanism}
% \noindent\textbf{Motivation.}
% Low probability is one of the prerequisites for creativity, yet it risks quality degradation. To prevent the model from drifting too far from the original distribution (e.g., producing out-of-distribution ``nonsense''), We implement two critically important pullback mechanisms. 
% \noindent\textbf{Anchor loss.}
% \noindent\textbf{MLLM validity checker.}
% each time we obtain an updated embedding $\tilde{e}$, we render an actual image via the diffusion decoder (Section~\ref{sec:ldm}). We then query a multimodal LLM to confirm that the generated image still matches the intended concept from prompt $y$. If it does, we continue the creative optimization; otherwise, we terminate that trial and restart sampling. This ensures that while we push the model toward embedding space regions of lower probability, the final outputs remain semantically meaningful.

\subsection{Pullback Mechanism}

% we need to add an argument to relate Pullback to Wundt curve. 

\label{PB}
As pointed out in Sec~\ref{motivation}, restricting the arousal potential is essential to avoid out-of-domain failure. Pushing the model too far from the original distribution risks degrading quality and semantic validity. To mitigate this, we introduce two pullback mechanisms that constraints the generated outputs to the intended concept: anchor loss and MLLM checker.

\noindent\textbf{Anchor Loss.} 
To ensure that the creative optimization does not cause the model to stray excessively, we enforce a CLIP-based~\cite{ramesh2022hierarchical} anchor loss: cosine similarity between the generated image embedding \(e\) and the text embedding of \(P_{pos}\). Formally,
\begin{equation}
\mathcal{L}_{\text{anchor}} = 1 - \frac{\langle e, \phi(P_{pos}) \rangle}{\|e\|\,\|\phi(P_{pos})\|}
\end{equation}
,where \(\phi\) denotes the text encoder. Anchor loss ensures that, despite the push toward low-probability (and potentially more creative) regions, the generated image remains semantically aligned with the subject described by \(P_{pos}\).

\noindent\textbf{Semantic Validity Checker.} 
 Complementing the anchor loss, we periodically validate the semantic consistency of our outputs using a multimodal large language model (MLLM). Every few optimization iterations, we generate an image embedding \(e\) from the optimized diffusion prior \(\epsilon_\theta\), then render an actual image using the diffusion decoder. We query the MLLM with the prompt ``Is this still an \(\{subject\}\)? Yes or No.'' If the MLLM confirms that the generated image corresponds to the intended concept, optimization proceeds; otherwise, the trial is terminated. This mechanism serves as an external semantic checkpoint, ensuring that our creative process does not yield out-of-distribution ``nonsense.'' We demonstrate that the MLLM is essential, even with an effective anchor loss, as evidenced in Sec.\ref{exp_pullback}.

Together, these pullback mechanisms balance the pursuit of creativity with the necessity of preserving quality and semantic fidelity in the generated images.

\subsection{Directionality}
\label{NC}
\noindent\textbf{Motivation.}
% Given the high dimensionality of the conceptual space, merely determining a suitable novelty level isn't sufficient. We must also identify the direction for sampling that yields compelling results. While the creative optimization steps in Sections~\ref{sec:creative-optimization} and~\ref{sec:semantic-check} can yield imaginative outputs, certain ``directions'' in the embedding space may lead to undesirable results—e.g., images that are semantically correct yet visually unappealing, or conversely visually striking but conceptually off-target. Because the trajectory of each creative trial is partly stochastic, the model can inadvertently converge on such unfavorable regions. Once identified, however, we can discourage revisiting these regions in future trials.
Despite the creative loss and pullback mechanisms ensures that generated images are both rare and semantically aligned with the intended concept, they do not guarantee outputs are compelling or interesting to humans. The inherently stochastic nature of the optimization process can drive the model toward specific regions that, while meeting these criteria, consistently yield unappealing or undesirable results. To address this challenge, we propose directionality control through negative cluster modeling.

% , merely enforcing low probability (creative loss) and  validity (pullback) conceptual spaces

\paragraph{Negative Cluster Modeling.}  
Suppose we observe a particular token embedding that consistently produces undesirable outputs, we generate negative image embeddings \(\{\tilde{e}_{neg\thicksim j}\}_{j=1}^{N}\). Similar to Section~\ref{PPS}, we project these embeddings into the same PCA-reduced space and fit a \emph{negative} multivariate Gaussian \(\hat{G}_{\text{neg}}(\tilde{e_{neg}})\). This distribution captures the ``unfavorable cluster'' we seek to avoid. In subsequent trials, We repel new samples from this unfavorable region by adding a penalty term to the creative loss:
\begin{equation}
    \mathcal{L}_\text{neg}(\tilde{e}) = -\,\alpha\,\log \hat{G}_{\text{neg}}(\tilde{e}),
\end{equation}
where \(\alpha\) is a strength scalar. 
% When combined with the original \emph{creative} objective from Sec.~\ref{CO}, t
The \emph{total} loss becomes:
\begin{equation}
    \mathcal{L} = \mathcal{L}_{\text{creative}}(\tilde{e}) + \mathcal{L}_\text{neg}(\tilde{e}) + \mathcal{L}_{\text{anchor}}.
\end{equation}
% Minimizing \(\mathcal{L}_{\text{creative}}\) still pushes \(\tilde{e}\) away from \emph{high}-probability regions in the baseline distribution \(\hat{G}(\tilde{e})\), but now also
This penalizes alignment with the negative cluster \(\hat{G}_{\text{neg}}(\tilde{e})\). Consequently, the model is guided to discover \emph{alternative} creative directions without getting trapped in regions known to produce unfavorable outcomes.

% By identifying these unfavorable clusters and incorporating a penalty term that discourages future sampling from these regions, we actively guide the model to explore alternative low-probability areas. This approach not only prevents the recurrence of suboptimal outputs but also fosters a more diverse and engaging exploration of the creative space.

%% file: sec/6_Implementation.tex
\section{Implementation Details}
\label{ID}
We operate on the official implementation of the Kandinsky 2.1 \cite{kandinsky2} text-to-image model and use its extended text encoder for input prompts, as recommended in~\cite{richardson2023conceptlab}. In the \textit{Prior Distribution Sampling} stage, we generate 5,000 image embeddings from the diffusion prior using \(P_{pos}\) with a diffusion step of 5 and a batch size of 500, a process that completes in \textit{\textcolor{Maroon}{less than one minute}}. The resulting embeddings, originally in \(\mathbb{R}^{768}\), are reduced via PCA to \(k=50\) dimensions, capturing the majority of the total variance. 

For the \textit{Creative Optimization} stage, training is performed on a single NVIDIA A100 GPU for up to 1,000 steps with a batch size of 1, using AdamW with a fixed learning rate of \(1\times10^{-4}\) to for both the token embeddings and the LoRA layers (rank = 10). Additionally, we query a multimodal LLM—employing either Janus-1.3B from DeepSeek-AI or LLaVA-Next—every 25 iterations for semantic validity check, automatically stopping a trial if out-of-domain results are detected. Empirically, creative and visually interesting outputs begin to appear within the first 50 steps, less than two minutes from the start.

%% file: sec/7_Experiments.tex
\section{Experiments}

% \subsection{Justify probabilistic creativity}
% we need evidence to prove the connection between \colorbox{red}{creativity, probability, and arousal potential.} \colorbox{red}{human eval on probability and arousal} \colorbox{red}{human eval on iteration and arousal}

% % In this section, we describe our experimental setup and evaluation strategy. We analyze both qualitative and quantitative aspects of our framework and compare it against baseline methods. 
% we now turn to prove the effectiveness of our method. 

% closely related to psychology. Secondary, can be put to appendix. 

\subsection{Visual Results.}

% \begin{figure}[ht]
%     \centering
%     % Placeholder for the actual figure
%     \includegraphics[width=\linewidth]{figures/result_22.pdf}
%     \caption{Generated creative images for alien and fruit.}
%     \label{fig:result_22}
% \end{figure}

Our method successfully generates highly creative and interesting images. Figure~\ref{hero_c} shows visual outputs for four subjects: \textit{building}, \textit{vehicle}, \textit{alien}, and \textit{fruit}. These examples illustrate that our method successfully generates highly novel, visually intriguing outputs across diverse categories. \textbf{We show extensively more visual results in the appendix.}
% The figure serves as qualitative evidence that our approach can produce images that deviate from conventional distributions, yielding novel and aesthetically engaging results.

\subsection{Evolution of the Generated Distribution}
\begin{figure}[ht]
    \centering
    % Placeholder for the actual figure
    \includegraphics[width=\linewidth]{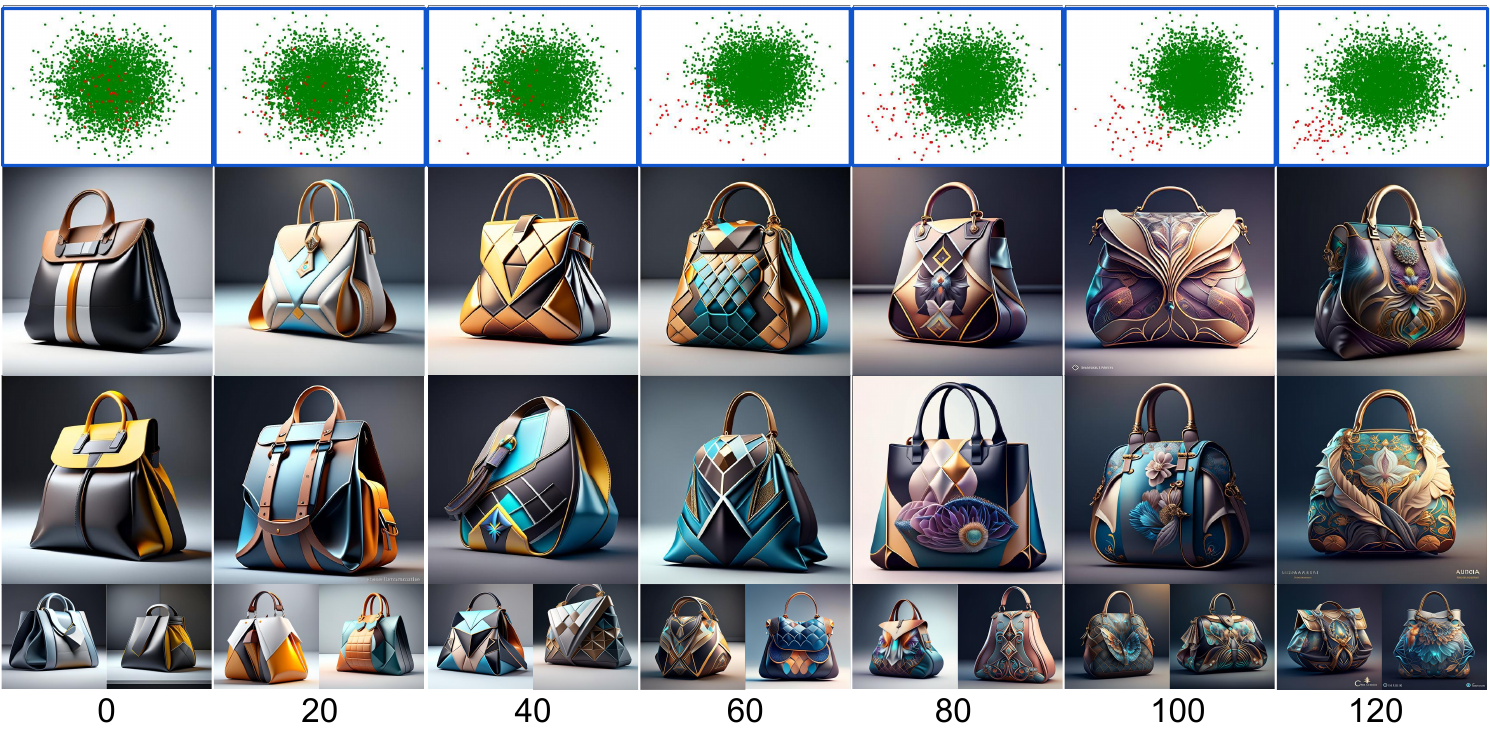}
    \caption{Visualization of the distribution of \(e\) shift over training iterations.  \textcolor{ForestGreen}{Green cluster} is the default distribution from prior sampling stage (Sec.\ref{PPS}). \textcolor{red}{Red cluster} is the current distribution. The generated images progressively move toward low-probability regions, resulting in more creative outputs over time. \textit{\textbf{(Zoom In)}}}
    \label{fig:result_3}
\end{figure}

To further validate our method, in Figure~\ref{fig:result_3}, we visualize how the distribution of the generated image embeddings \(e\) evolves during training. As the optimization proceeds, the distribution of \(e\) progressively shifts toward the boundary, moving into lower-probability regions. Correspondingly, the generated images become increasingly creative over time. This confirms the effectiveness of our method in pushing the model toward novel regions, and also demonstrates a clear correlation between the distribution tailward-shifting and the creativity of the outputs.

\subsection{Human Evaluation of Arousal Potential}
In Sec.\ref{motivation}, we talk about the Wundt curve~\cite{Berlyne1971,wundt1874grundzuge}: the connection between arousal potential, creativity, and probability. In this experiment, we show experiments on subject \textit{alien} to validate this concept. For this purpose, we \textit{\textcolor{BlueViolet}{temporarily disable the pullback mechanisms}} and run experiment trial with 5 different training seed. We collect human evaluation ratings of the creativity of generated images on a scale from 0 to 5 (Detailed in the appendix). The resulting figure (see Figure~\ref{fig:arousal_1}) reveals a clear pattern: creativity score progressively rise from boring to interesting (around 3) and very interesting (around 4), before eventually declining to 0 as the model overshoots the optimal arousal range \footnote{Default output is used as an example image in the survey and its creative score is set to 1 (boring). Score 0 means out-of-domain.}. This trajectory mirrors the expected arousal potential curve, validating the underlying theoretical framework, and proves our method can effectively explore the curve \footnote{We explain Seed 5 in the appendix.}. 

\begin{figure}[ht]
    \centering
    % Placeholder for the actual figure
    \includegraphics[width=\linewidth]{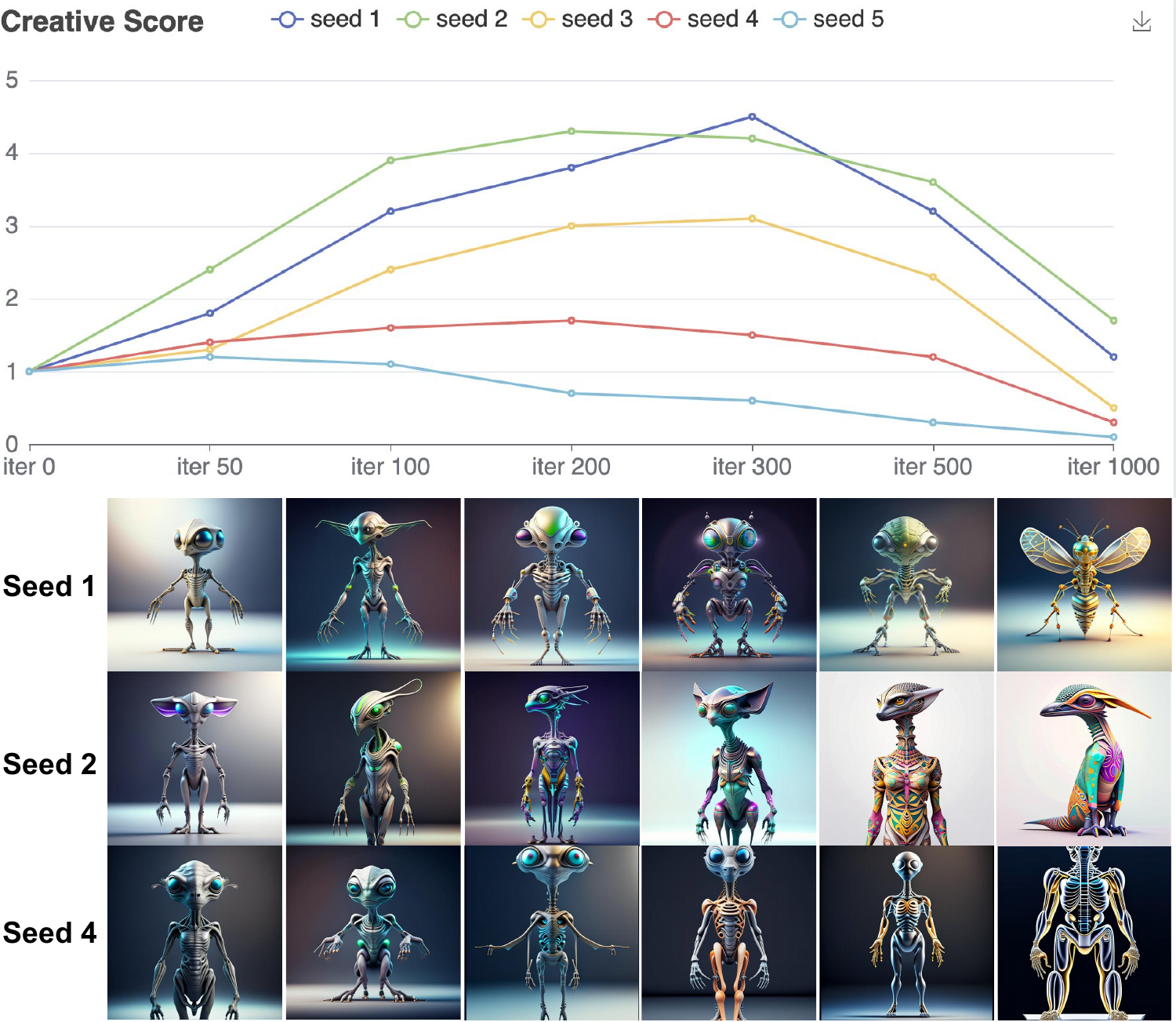}
    \caption{User-rated creativity scores over training iterations for the subject "alien." The observed pattern follows the arousal potential curve, demonstrating how our method \textcolor{BlueViolet}{(pullback disabled)} initially enhances creativity effectively before overshooting. We show visual samples for 3 seeds, omitting default ones (iter 0).}
    \label{fig:arousal_1}
\end{figure}

\begin{figure}[ht]
    \centering
    % Placeholder for the actual figure
    \includegraphics[width=\linewidth]{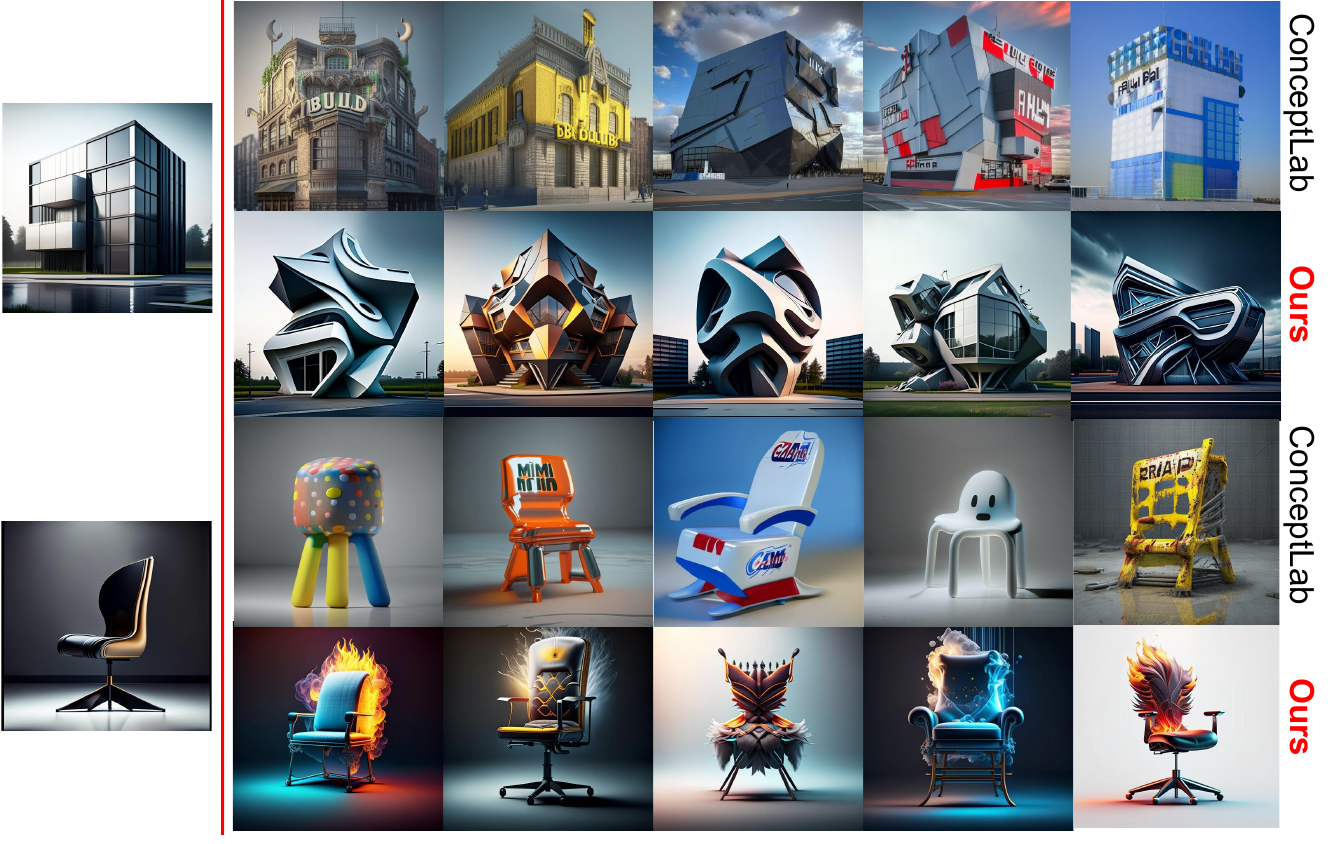}
    \caption{Generated creative images for building and chair.}
    \label{fig:result_compare}
\end{figure}

% We present visual examples of generated images under various text prompts, comparing our model's outputs to baseline models. We emphasize creativity, distinctiveness, and fidelity.

% \subsection{Quantitative Metrics}
% Compare with conceptlab, human evaluation. 
% We introduce objective measures to evaluate our model, including:
% \begin{itemize}
%     \item \textbf{Human Evaluation} We conduct a human evaluation where participants rate generated images on two aspects: \textit{creativity} (novelty and uniqueness) and \textit{semantic accuracy} (how well the image aligns with the given text prompt). The results provide insights into user perception of our model.
    % \item \textbf{Embedding-Space Diversity:} Measuring dispersion in the PCA-reduced space.
    % \item \textbf{Novelty Scores:} Estimating uniqueness based on distance to prior samples.
    % \item \textbf{Fidelity Metrics:} Evaluating perceptual quality using CLIP similarity.
    % \item \textbf{Diversity Metrics:} Evaluating diversity using CLIP similarity.    
% \end{itemize}
% \colorbox{red}{ongoing: human score change with iteration.}

% \colorbox{red}{}{red}{??? cite a paper about how FID is not effective. }

\subsection{Compare with Baseline}
% This section compares ours with the baseline: ConceptLab.

\subsubsection{Quality and Speed.}  
Figure~\ref{fig:result_compare} compares our method against ConceptLab on the subjects of buildings and chairs. Our approach produces significantly more creative and conceptually rich samples, while the baseline tends to generate images that are either overly conventional or lack creative expression in a desired way. 
These qualitative observations are further supported by quantitative human evaluations presented in Sec.~\ref{sec:human_evaluation}.

\begin{figure}[ht]
% \vskip 0.2in
\begin{center}
\includegraphics[width=1\linewidth]{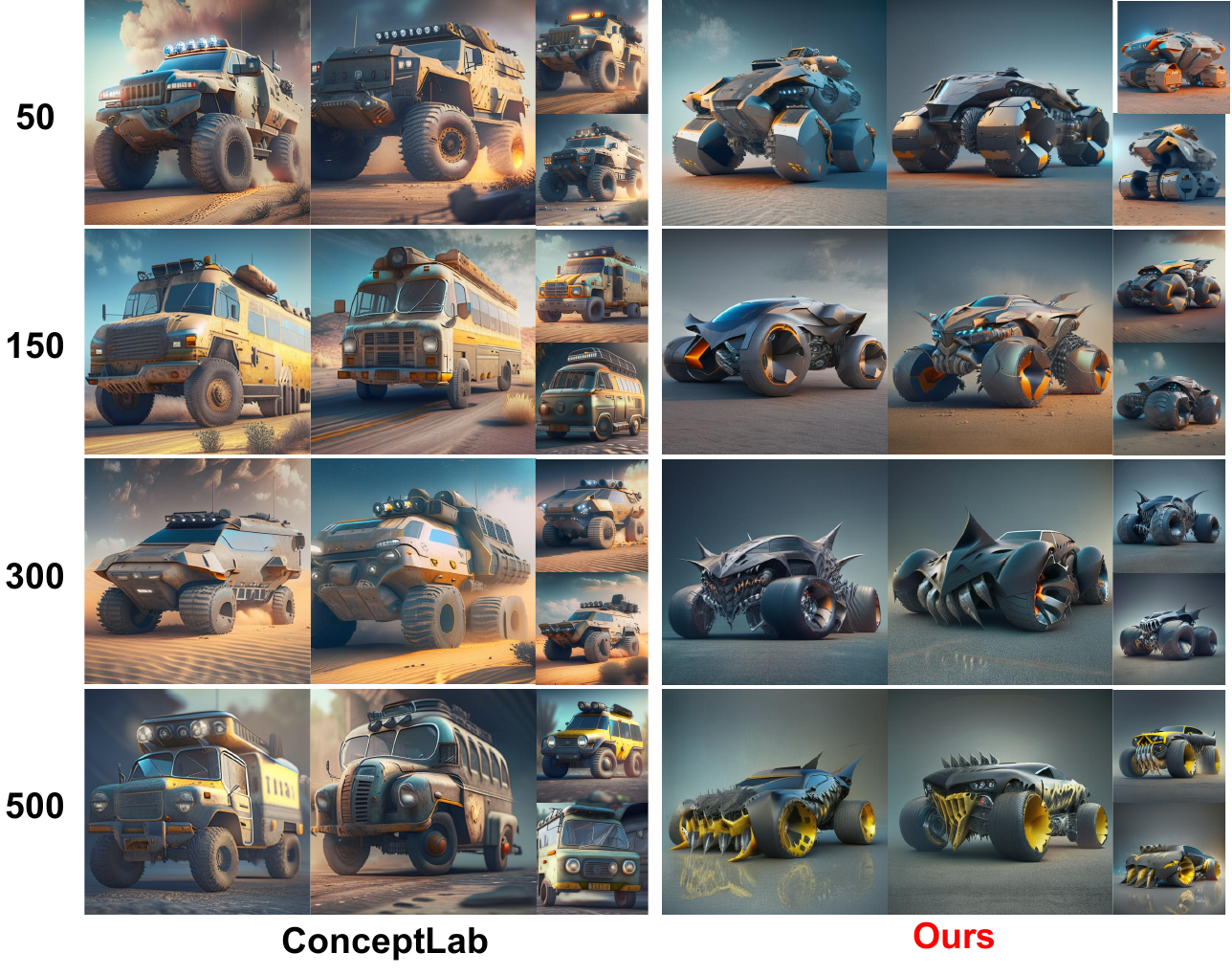}
\caption{Comparison of creative vehicle generations at different iterations (50, 150, 300, 500). ConceptLab (left) gradually explores subcategories before reaching creative designs. Our method (right) is not only faster, but also much more creative.}
\label{Compare_1}
\end{center}
% \vskip -0.2in
\end{figure}

% \subsubsection{Speed Comparison}  
To validate the speed and efficiency of our method, we compare how generated images evolve during training on subject \textit{vehicle}. In Figure~\ref{Compare_1}, the left panel shows ConceptLab’s outputs, where creative samples emerge after 300 iterations: as ConceptLab achieves creativity by avoiding subclasses, it initially produces subcategory-specific results (e.g., jeep, bus) before being forced to more creative outcomes. In contrast, our method, operating directly on the distribution and probability space, yields creative outputs from the very beginning. This accelerated convergence not only highlights the efficiency of our approach but also aligns with our intuition: operating on distribution bypasses the need for explicit subcategory exclusion, allowing for faster exploration of novel, low-probability regions in the embedding space. \footnote{Prior distribution sampling takes less than 1 minute. See Sec.\ref{ID}.}

\subsubsection{Human Evaluation}
\label{sec:human_evaluation}
\begin{figure}[ht]
    \centering
    % Replace 'placeholder_path/human_eval.pdf' with the actual figure path
    \includegraphics[width=1.0\linewidth]{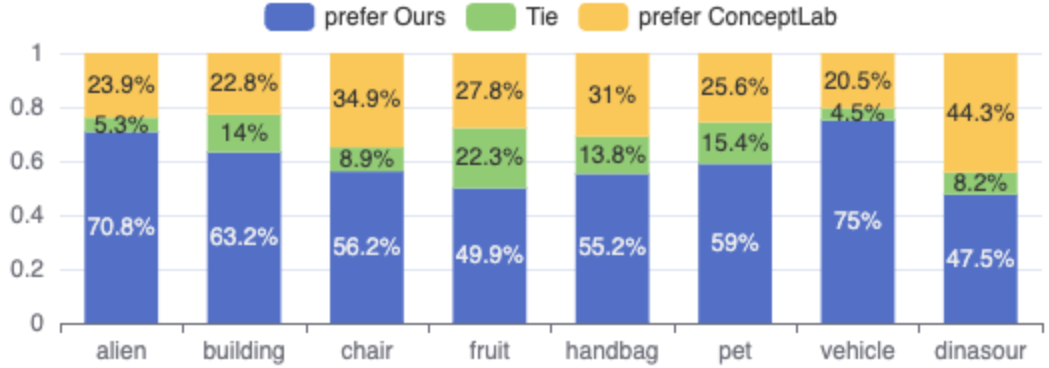}
    \caption{Human evaluation comparing the creativity of outputs from both methods. Our model stably outperforms the baseline.}
    \label{fig:human_eval}
\end{figure}
To quantitatively evaluate the quality of outputs from both methods, we conducted Human Evaluations across eight subjects. Participants were shown paired images (three from each method) and asked which they found more creative. Figure~\ref{fig:human_eval} shows that a clear majority preferred our approach for all categories. For example, \textbf{70.8\%} of participants favored our \emph{alien} images, and \textbf{75\%} for our \emph{vehicles}.
% : \emph{alien, building, chair, fruit, handbag, pet, vehicle,} and \emph{dinosaur}
 % preferred ours over ConceptLab
% , and \textbf{20.7\%} found no difference. Other categories such as \emph{chair}, \emph{fruit}, and \emph{vehicle} also saw a higher preference for our outputs. The \emph{dinosaur} category showed a closer split, suggesting that both methods can produce compelling yet distinct creative styles. 
These results indicate that our method consistently yields more creative, novel, and engaging outputs.

% \subsection{Experimental Setup}
% This subsection details the technical environment, including computational resources and hyperparameter settings. The setup follows the specifications outlined in Section~3.4 to ensure reproducibility.

\subsection{Ablation Studies}

We conduct extensive experiments to better demonstrate the effectiveness of each of our contribution components. Below, we describe the details of each ablation study.

% simple baseline just like conceptlab
\subsubsection{Invalidate Direct Image Embedding Optimization}

\begin{figure}[ht]
% \vskip 0.2in
\begin{center}
\includegraphics[width=1\linewidth]{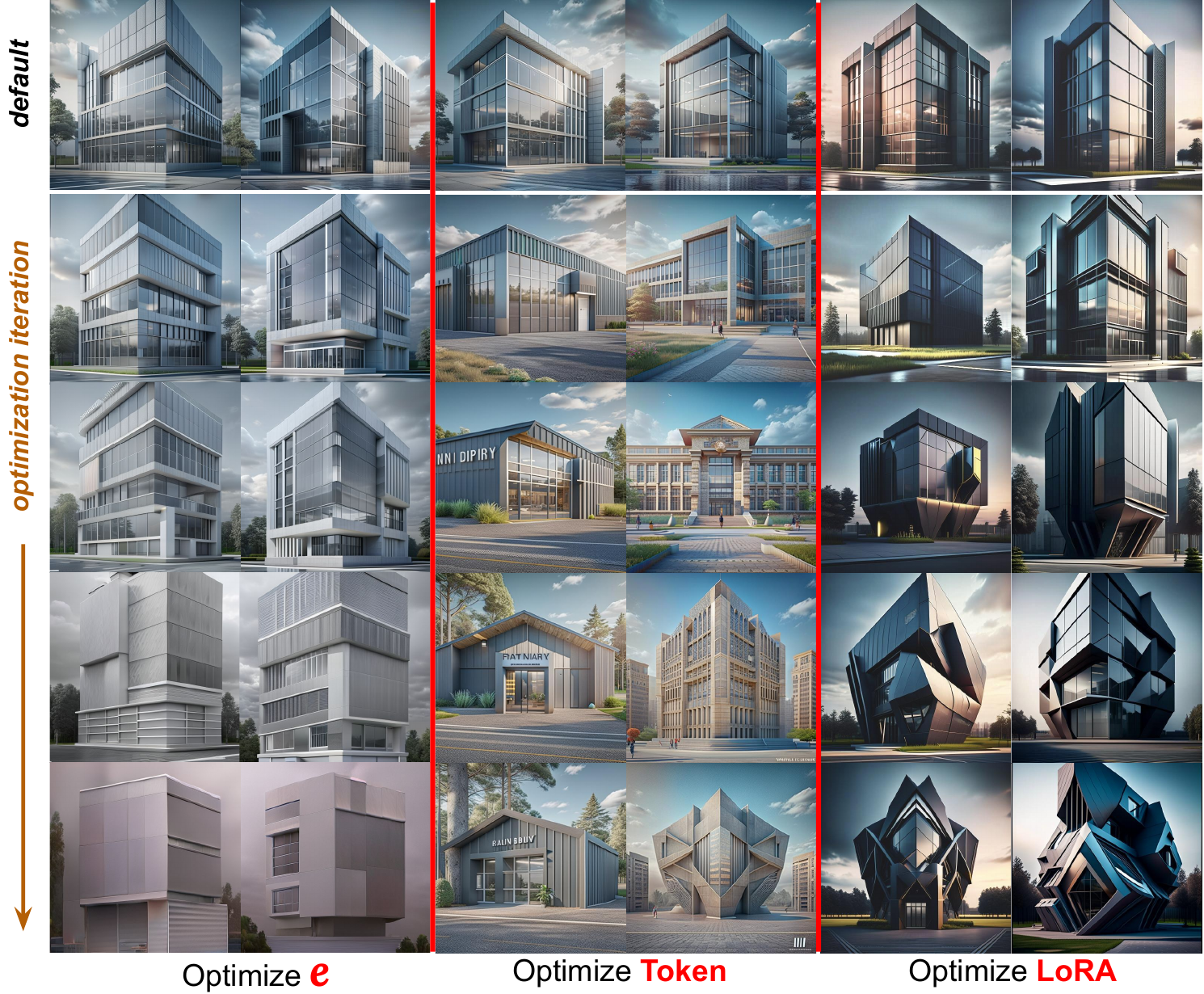}
\caption{Comparison of optimization strategies. 
% Seeds are fixed so we can track the same image. 
    \emph{Optimize $e$} (left column) quickly leads to degraded outputs. 
    \emph{Optimize subject Token} (middle column) and \emph{Optimize LoRA} (right column) both yield creative and valid results.}
\label{lora}
\end{center}
% \vskip -0.2in
\end{figure}

To investigate the role of optimizing Token/LoRA in our method, we compare three strategies for incorporating our creative loss: (1) directly optimizing the image embedding $e$, (2) optimizing a word embedding, and (3) optimizing the LoRA parameters of the diffusion prior. 

Figure~\ref{lora} shows how the images evolve over 200 iterations for each approach. Directly optimizing $e$ (left column) causes the output to degrade quickly, suggesting that it's prone to easily slipping into bad-quality images since it lacks quality constraints. By contrast, optimizing the subject token (middle column) and the LoRA parameters (right column) both yield creative yet coherent transformations. Notably, LoRA optimization excels at adding complex architectural details and variations, as it leverages a broader parameter space to search in. We explain this in detail in the next subsection.
% Overall, this comparison underscores the importance of selecting an appropriate optimization target when pushing the model toward creative regions.
% , with the latter  LoRA-based optimization , demonstrating its flexibility

% One can imagine an approach where we can generate a large number of samples in a CLIP space to model the distribution, then define a loss function to minimize equation 1 directly in that space. This method would be prone to easily slipping into bad quality images since it lacks constraints on the samples that insure its quality.

\subsubsection{Subject Token vs. Adjective Token vs. LoRA}
\begin{figure}[ht]
% \vskip 0.2in
\begin{center}
\includegraphics[width=1\linewidth]{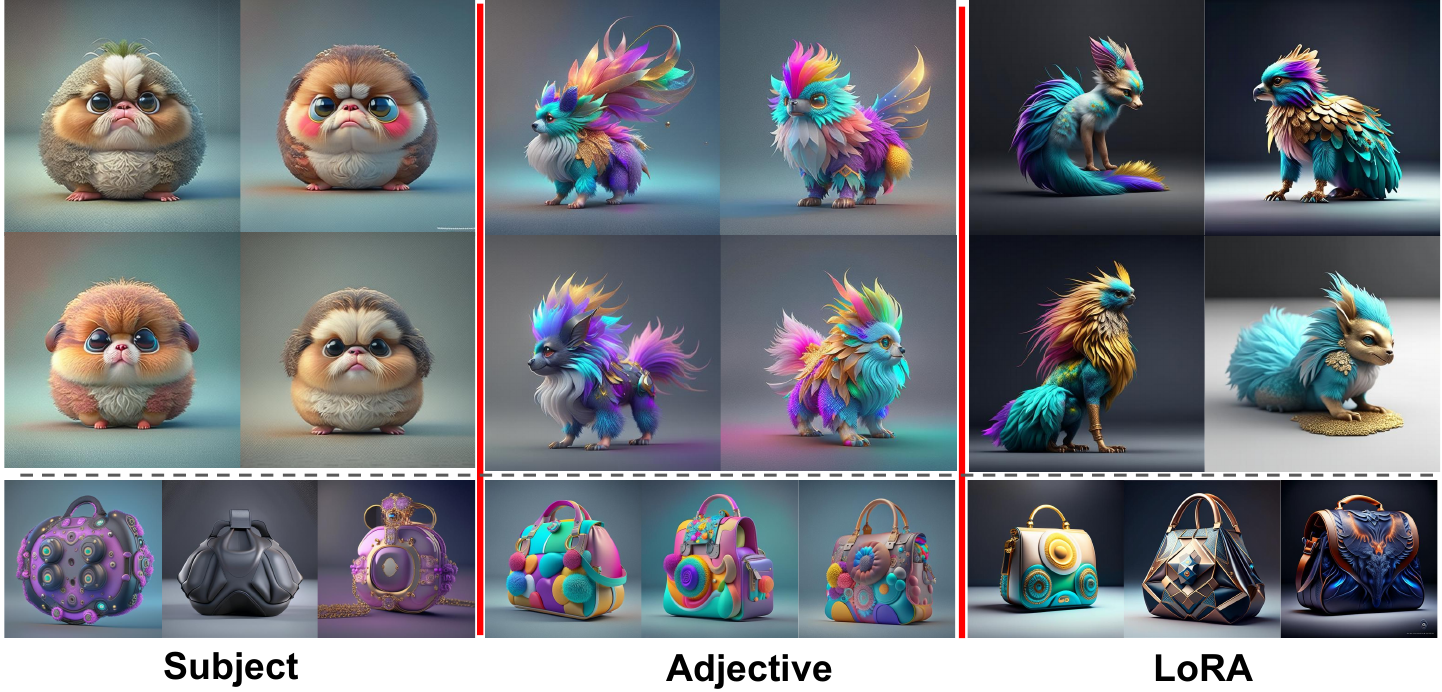}
\caption{Comparison of three conceptual spaces. For \emph{handbags}, we show 3 trials with different training seeds. Optimizing the subject token sometimes fails to discover creative outputs, especially for pets, while optimizing an adjective token can lead to consistently more creative images. LoRA fine-tuning offers a more diverse exploration of creative possibilities, especially for handbags, at the cost of being more resource-demanding.}
\label{fig:subject_vs_adjective_vs_lora}
\end{center}
% \vskip -0.2in
\end{figure}

To further understand the role of conceptual space, we explore multiple options for creative optimization: (1) optimizing the \emph{subject token}, (2) optimizing an \emph{adjective token} \footnote{An example prompt is: Professional high-quality photo of a \textcolor{red}{creative} handbag. photorealistic, 4k, HQ. The adjective "creative" is optimized.}, and (3) optimizing LoRA.
% The former replaces the literal subject word (e.g., ``\texttt{handbag}'' or ``\texttt{pet}'') with a learned embedding, while the latter modifies an adjective slot (e.g., ``\texttt{creative}'') in the prompt \emph{without} altering the subject itself. This allows us to leverage the strong constraints of language prompts to maintain semantic coherence. Finally, we compare both to fine-tuning LoRA parameters in the diffusion prior.
Figure~\ref{fig:subject_vs_adjective_vs_lora} shows results for \emph{pets} and \emph{handbags} under each approach. We observe that \textbf{optimizing the subject token} works sub-optimally for \emph{handbags} and struggles for \emph{pets}, either distorting subject identity or lacking creativity. In contrast, \textbf{optimizing an adjective token} (e.g., ``\texttt{creative}'') consistently produces highly creative outcomes. However, this strategy tends to converge on vivid or colorful as an efficient way to deviate from the distribution, potentially limiting variety. By comparison, \textbf{optimizing LoRA} parameters yields a broader range of creative transformations that do not rely exclusively on color shifts. In particular, LoRA fine-tuning can discover more structural or stylistic alterations while retaining the underlying subject identity. Overall, these findings highlight the advantages of exploring different parameter spaces for creativity: subject tokens, adjective tokens, and LoRA each offer distinct trade-offs between visual diversity, semantic fidelity, and ease of optimization.

% (subject token, adjective token, LoRA, full model, multiple adjectives, prompt, etc.) We compare our LoRA-based fine-tuning approach against full fine-tuning of the diffusion prior. Our analysis highlights differences in computational efficiency, stability, and generation quality.

\subsubsection{Directionality}
\begin{figure}[ht]
\vskip 0.2in
\begin{center}
\includegraphics[width=1\linewidth]{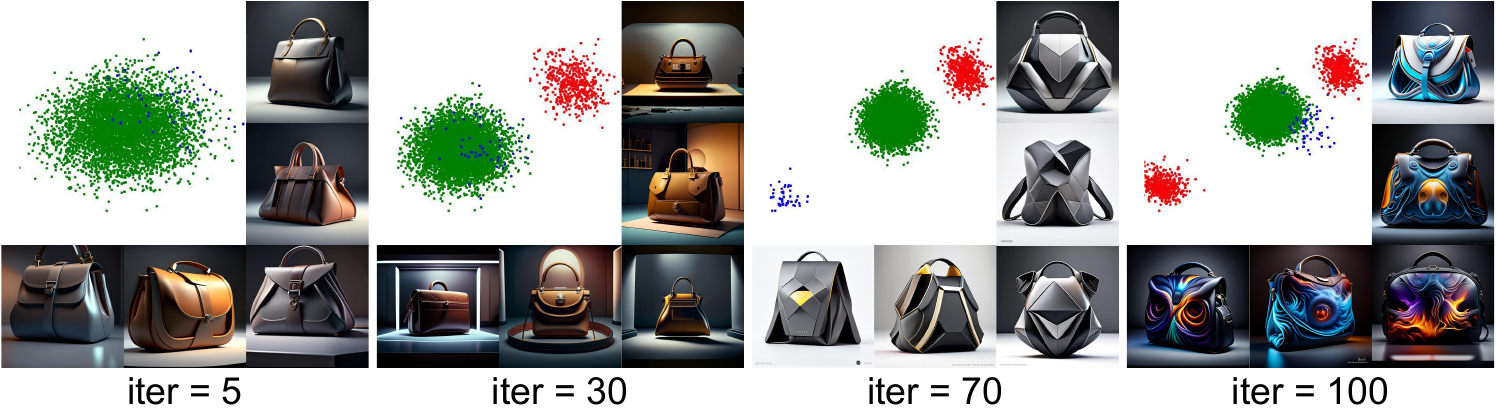}
\caption{Direction control for creative handbag. The model initially reduces probability with visually unappealing outputs. Labeling these as negative clusters steers the model toward alternatives. (We project \(e\) to 2D only for visualization.)}
\label{fig:negative_cluster}
\end{center}
\vskip -0.2in
\end{figure}

As discussed in Sec.~\ref{NC},
% Although sampling from low-probability regions can yield highly creative outputs, it does not guarantee favorable or high-quality results. Some directions may reduce probability by altering irrelevant factors (e.g., backgrounds) or producing visually unappealing styles. To address this, 
we label visually unappealing regions as \emph{negative clusters} and penalize the optimized distribution aligns with them, forcing exploration of alternative directions. Figure~\ref{fig:negative_cluster} illustrates this process for the task of generating creative \emph{handbags}. Initially, the model seeks to lower probability by modifying the background in undesirable ways. Once identified, these embeddings are used to draw a negative cluster, steering the model to avoid that direction. The system then tries a second approach, producing an unfavorable aesthetic style; hence, it too becomes a negative cluster. Finally, the model settles on a new, more appealing style---achieving creativity through novel shapes and color patterns. This outcome highlights the importance of \emph{directionality control}, ensuring that probability tail exploration remains aligned with user preferences.

% \begin{figure}[ht]
% % \vskip 0.2in
% \begin{center}
% \includegraphics[width=1\linewidth]{figures/result_2.pdf}
% \caption{Negative cluster avoidance in creative handbag generation. The model initially reduces probability by altering backgrounds, yielding visually unappealing outputs. Labeling these as negative clusters steers the model toward alternative low-probability directions, eventually producing more appealing and innovative designs. (Zoom In)}
% \label{fig:negative_cluster}
% \end{center}
% % \vskip -0.2in
% \end{figure}

\subsubsection{Pullback Mechanism}
\label{exp_pullback}
\begin{figure}[ht]
\vskip 0.2in
\begin{center}
\includegraphics[width=1\linewidth]{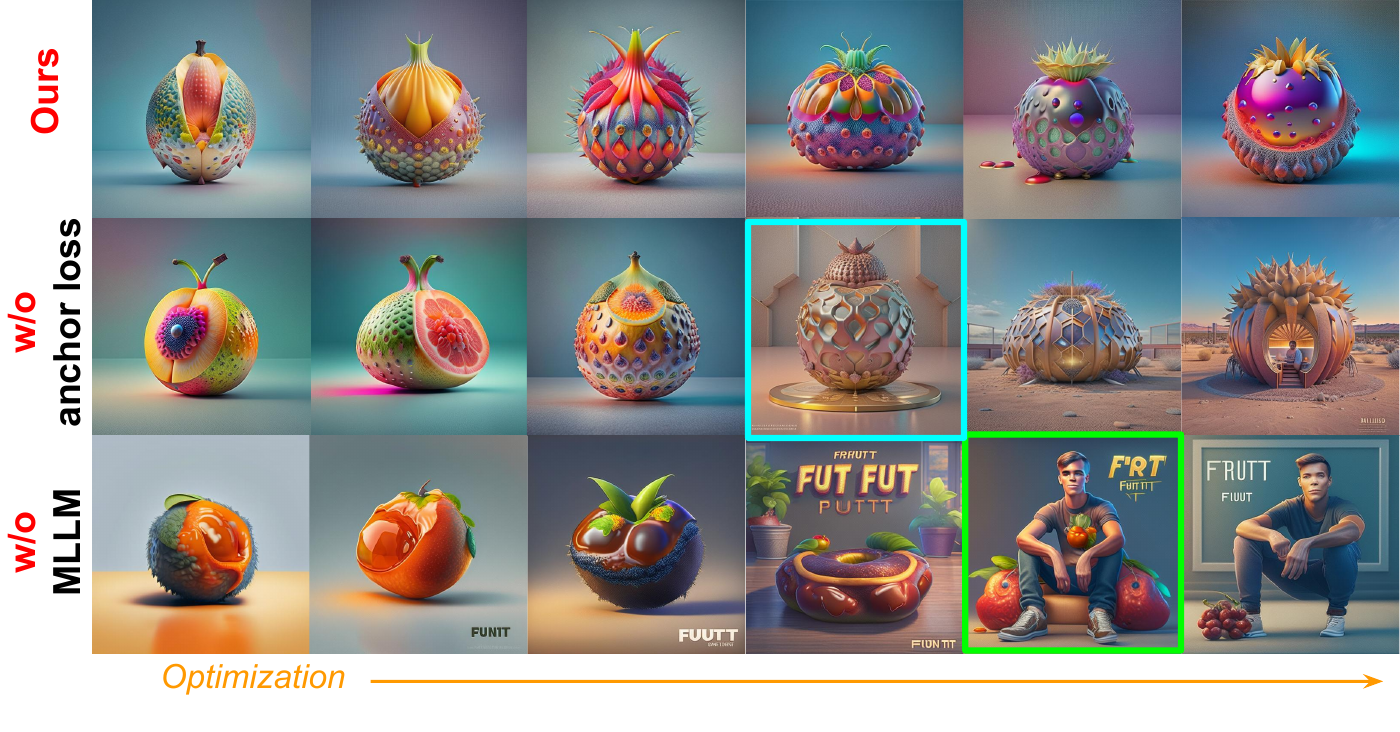}
\caption{Ablation on the pullback mechanism for creative fruits. 
% This demonstrates the importance of both pullback components in ensuring semantic validity when promoting creativity.
}
\label{fig:pullback_fruit}
\end{center}
\vskip -0.2in
\end{figure}

Figure~\ref{fig:pullback_fruit} illustrates the vital impact of our two pullback components: anchor loss and a Semantic (MLLM) Validity Checker. When generating creative fruits, our complete method (top row) balances creativity and semantic fidelity. 
Removing the anchor loss (middle row) gradually drifts out-of-domain, starting from \textcolor{cyan}{Blue Box}.
% Removing anchor loss (middle row) causes the system to  (e.g., producing building-like structures), 
Removing the MLLM Checker \emph{Bottom row} results in deceptive adherence to the ``\texttt{fruit}'' by inserting small fruit motifs in the image but morphing the main subject into a human figure. Semantic checker would have terminates it on \textcolor{ForestGreen}{green box}. This highlights that CLIP-based constraint is subject to adversarial attack, reinforcing the necessity of both pullback mechanisms to maintain in-domain creativity.
% alone can be adversarially circumvented, 

% \textcolor{red}{optional: show how fast it goes out of domain. }

%% file: sec/8_Discussion.tex
\section{Discussion}
\label{future}
\noindent\textbf{Novelty.} To our best knowledge, this is the first probabilistic creativity framework for T2I generation.  We differ fundamentally from ConceptLab in that it does not rely on the exclusion of known subclasses to generate creative outputs. ConceptLab operates by first visiting and then excluding intermediate subcategory representations, which can be ineffective for subjects lacking clear subclasses (e.g., chairs and aliens) or inefficient for subjects with well-defined subclasses. By steering the model toward probability tails without any dependency on subclass structure, our technique accesses creativity much more efficiently.

\noindent\textbf{Extension, limitation and more.} 
We chose Kandinsky for demonstration due to its lightweight prior model, which enables rapid stage-one prior sampling. However, our method can also be applied to other frameworks, for example \textit{\textbf{Hyper-SD}}. Due to page limits, we show the model diagram, results, evaluations and limitations in the \textcolor{violet}{appendix}. 
% Our work demonstrates that exploring low-probability regions in the intermediate image embedding space can significantly enhance creative outputs. However, many state-of-the-art text-to-image models, such as Stable Diffusion 3 and Flux, do not explicitly provide access to such an intermediate space. Future work should investigate how to adapt our approach to these models.
% \footnote{Due to space limitations, additional experiments and analyses have been deferred to the supplementary material.
% , which includes: the influence of random seeds on the creative trajectory, an analysis on the influence of PCA dimensionality, effect of replacing Gaussian with Kernel Density Estimation, and more visual/quantitative results. 

% seed beam search, secondary, move to discussion (appendix)
% future work latent space in other diffusion models.
% \subsubsection{Effect of PCA Dimensionality}. 
% also KDE, secondary. put to the appendix.  We investigate the impact of the number of principal components \( k \) on both creativity and fidelity. We compare model performance for \( k \in \{20, 50, 100\} \) and analyze how dimensionality reduction affects probability estimation.
% secondary, move to discussion (appendix)

%% file: sec/9_Conclusion.tex
\section{Conclusion}
In this work, we presented a principled approach to fostering creativity in text-to-image generative models by explicitly targeting probability tails of the generated image embedding distribution. Building on a latent diffusion framework, we introduced a creative loss, pullback mechanism, and direction control to drive the model toward novel yet semantically valid outputs. Through comprehensive experiments and ablation studies, we demonstrated the efficiency and effectiveness of our method. We believe that this work initiates new directions for computational creativity, serving as a solid first step toward more expressive, flexible, and creative generative AI systems.

%% file: sec/X_suppl.tex
\noindent\textbf{Supplementary Material}

\section{More visual evidence. }
\begin{figure}[ht]
\begin{center}
\includegraphics[width=0.9\linewidth]{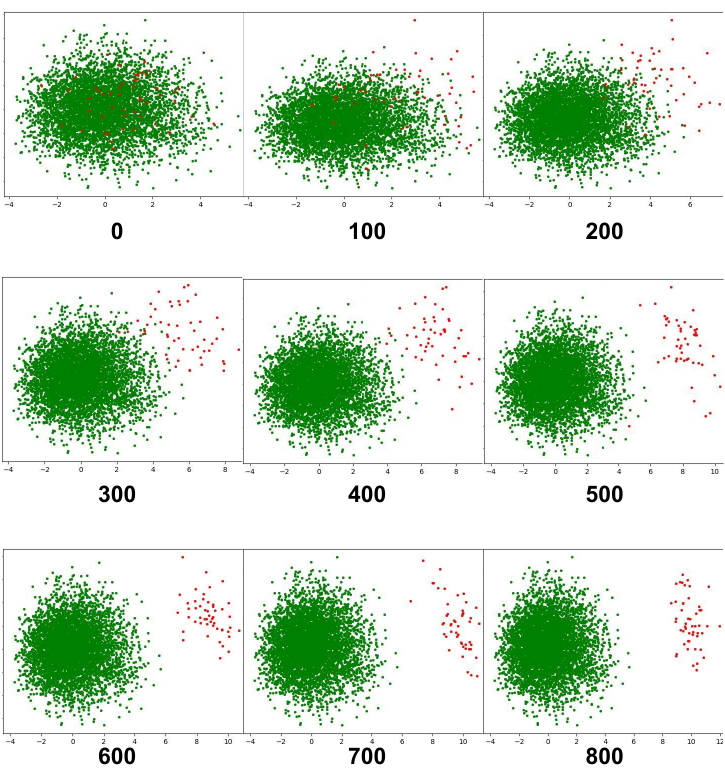}
\caption{Distribution Evolution for alien experiments for the first 800 iterations. The green cluster is the prior distribution, and the red cluster is the optimized distribution during creative tuning. Our method effectively drives the distribution towards its tail, forcing it to explore rare and valid novel designs.}
\label{humaneval_22}
\end{center}
\end{figure}

In Fig. \ref{humaneval_22}, we show one example of how the distribution evolves overtime. 

We also attach much more visual evidence, including additional qualitative results across various subjects, visualizations of the optimized distribution, and further comparisons with baseline methods to reinforce the effectiveness of our approach. Please find them at the end of this appendix.

%%%%%%%%%%%%%%%%%

\begin{figure*}[ht]
\begin{center}
\includegraphics[width=0.9\linewidth]{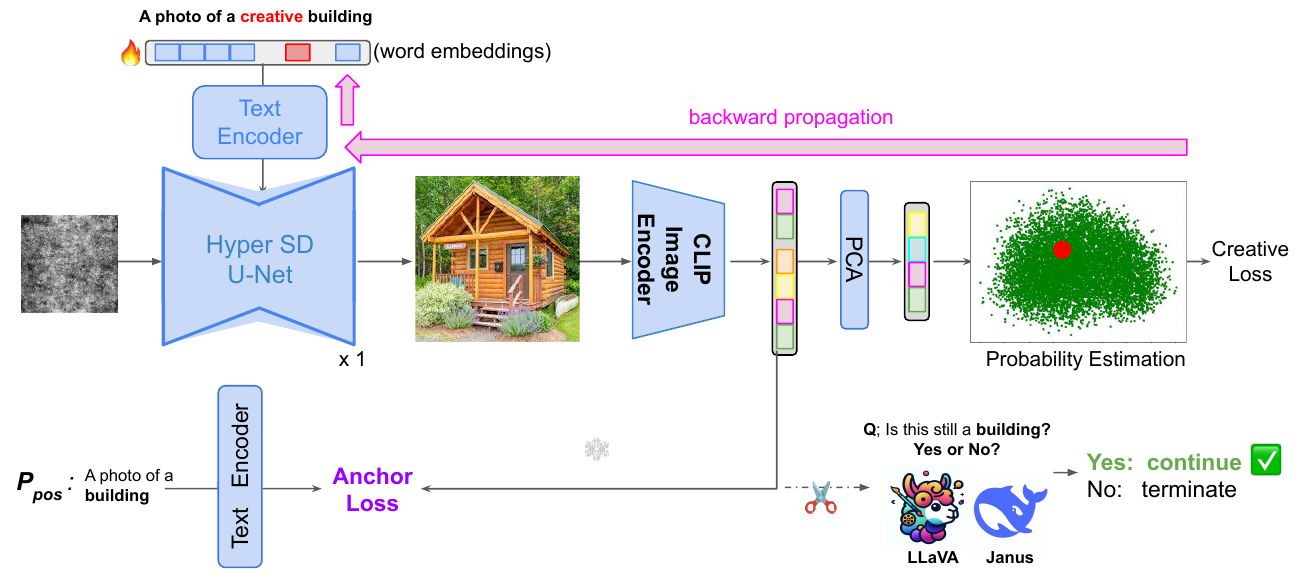}
\caption{Extend our method to Stable Diffusion.}
\label{SD_extension}
\end{center}
\end{figure*}

\begin{figure}[ht]
\begin{center}
\includegraphics[width=1.0\linewidth]{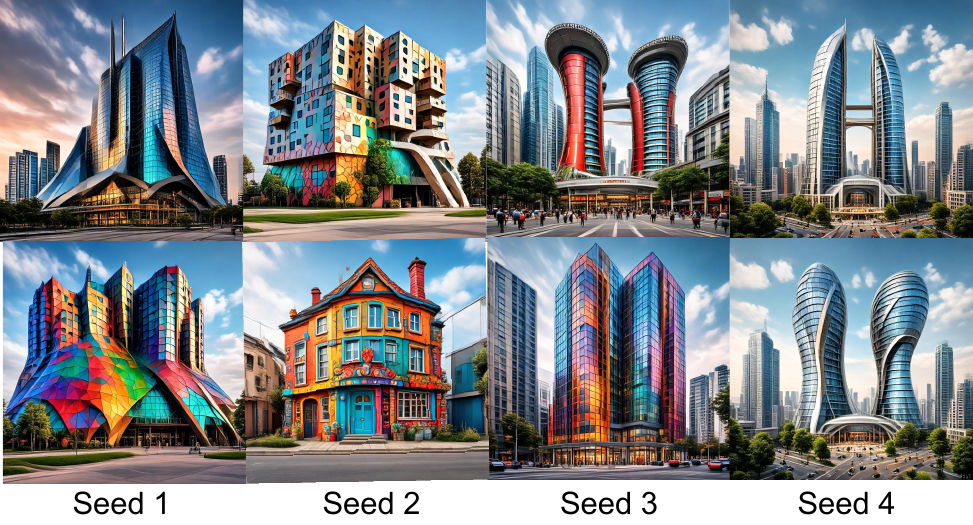}
\caption{Results of creative Hyper-SD. Creative Building generations using 4 different seeds.}
\label{SD_building}
\end{center}
\end{figure}

\section{Implementation Details.}
\subsection{PCA dimension.} We choose to apply PCA reduction and use a dimension of 50 for our creative generation task based on the following considerations:
\begin{itemize}
    \item Empirically, reducing the image embedding from 768 to 50 dimensions retains over 95\% of the variance, ensuring that most of the important information is preserved.
    \item The reduced 50-dimensional space is compact yet rich enough to effectively explore creative variations, allowing our optimization process to focus on meaningful differences in the embedding space.
    \item Lower dimensionality reduces computational complexity and noise, facilitating more stable density estimation and making the Gaussian approximation more effective for our creative generation task.
    \item PCA helps to simplify the high-dimensional embedding space, enabling more efficient identification of low-probability regions while filtering out redundant or less informative components.
\end{itemize}

\subsection{Probability estimation Using a Gaussian Fit.}
Our decision to fit a Gaussian distribution to the image embeddings sampled from the diffusion prior is supported by several factors intrinsic to the diffusion model framework:

\begin{enumerate}
\item \textbf{Natural Emergence of Gaussianity:} Diffusion models inherently rely on the gradual injection and removal of Gaussian noise. As a result, the intermediate image embeddings produced by the diffusion prior tend to exhibit Gaussian-like behavior, especially as we observed in Kandinsky 2.1\cite{kandinsky2}. This makes a multivariate Gaussian a natural choice for modeling the underlying distribution.
    
    \item \textbf{Computational Efficiency and Simplicity:} Fitting a Gaussian is computationally efficient and offers a closed-form solution for density estimation. This simplicity facilitates the direct calculation of log-likelihoods, which is crucial for our creative loss.
    
    \item \textbf{Sufficient Approximation for Creative Exploration:} Our creative generation task focuses on steering the model towards low-probability regions. Empirically, a Gaussian model provides a robust approximation of the diffusion prior's embedding space, enabling us to effectively identify and explore the tail regions where novel and creative outputs are more likely to emerge.
\end{enumerate}

\paragraph{Alternative Density Estimation via KDE.} We also explored using kernel density estimation (KDE) as an alternative method for modeling the distribution of image embeddings. Although KDE offers a flexible, non-parametric density estimation, it requires reducing the PCA dimensionality to a much smaller value to achieve reliable performance. This substantial reduction in dimensionality can discard important variance in the embedding space, thereby limiting the diversity and potential for creative exploration. Consequently, we chose the Gaussian approximation, which maintains a balance between computational efficiency and expressive capacity, ensuring that our method preserves sufficient creative diversity.

\subsection{Impact of Training Seed.} The choice of training seed significantly influences the creative trajectory of our model. Different seeds lead to diverse initializations and sampling paths, thereby steering the model toward distinct regions in the embedding space. This variability means that a simple change in seed can yield a completely new set of creative samples. However, it is also natural for some seeds to produce unfavorable or out-of-domain results, see Seed 5 in Figure~\ref{fig:arousal_1}, a reflection of the inherent unpredictability in creative generation. This observation further justifies our use of dimensionality control measures, such as negative cluster avoidance, to steer the model away from undesirable outcomes and maintain semantic coherence.

\subsection{Anchor loss.} A key challenge is that the creative loss and anchor loss yield substantially different gradient magnitudes: the creative loss produces very large gradients that quickly push the distribution toward the boundary, whereas the anchor loss yields much smaller gradients that require many more optimization steps to have an effect. Although gradient clipping can alleviate this imbalance, it cannot fully resolve it, and finding a fixed reweighting factor is difficult because the optimal weight may vary with experiments and subjects. To address this, we adopt a dynamic strategy: 
\begin{itemize}
\item in each iteration, we sample a seed to generate an image embedding and compute both the creative loss and the anchor loss. 
\item If the anchor loss is below a predetermined threshold, we set the overall loss equal to the creative loss and sample a new seed in the next iteration. 
\item If, on the other hand, the anchor loss exceeds the threshold, we set the overall loss to the anchor loss and \textit{\textcolor{Maroon}{keep optimizing using the same seed}} until we pull it back. 
\end{itemize}

\noindent This approach ensures that we persist in optimizing the same sample until it is successfully pulled back within acceptable limits, thereby balancing the aggressive influence of the creative loss with the stabilizing pull of the anchor loss.

\section{Extension to Other Frameworks.}
Our method focuses on a probability approach for creative image generation, and can be extended to frameworks other than Kandinsky. For example, in Fig. \ref{SD_extension}, we show our extension to a Hyper-SD, a Stable Diffusion distillation method. We show its results on creative building generation in Fig. \ref{SD_building}.

%%%%%%%%%%%%%%%%%
\section{Large scale Visual results. }
Starting from this point, we show large amount of visual results.

\noindent\textcolor{red}{\textbf{See Next Page.}}

\clearpage

\begin{figure*}[ht]
\begin{center}
\includegraphics[width=0.85\linewidth]{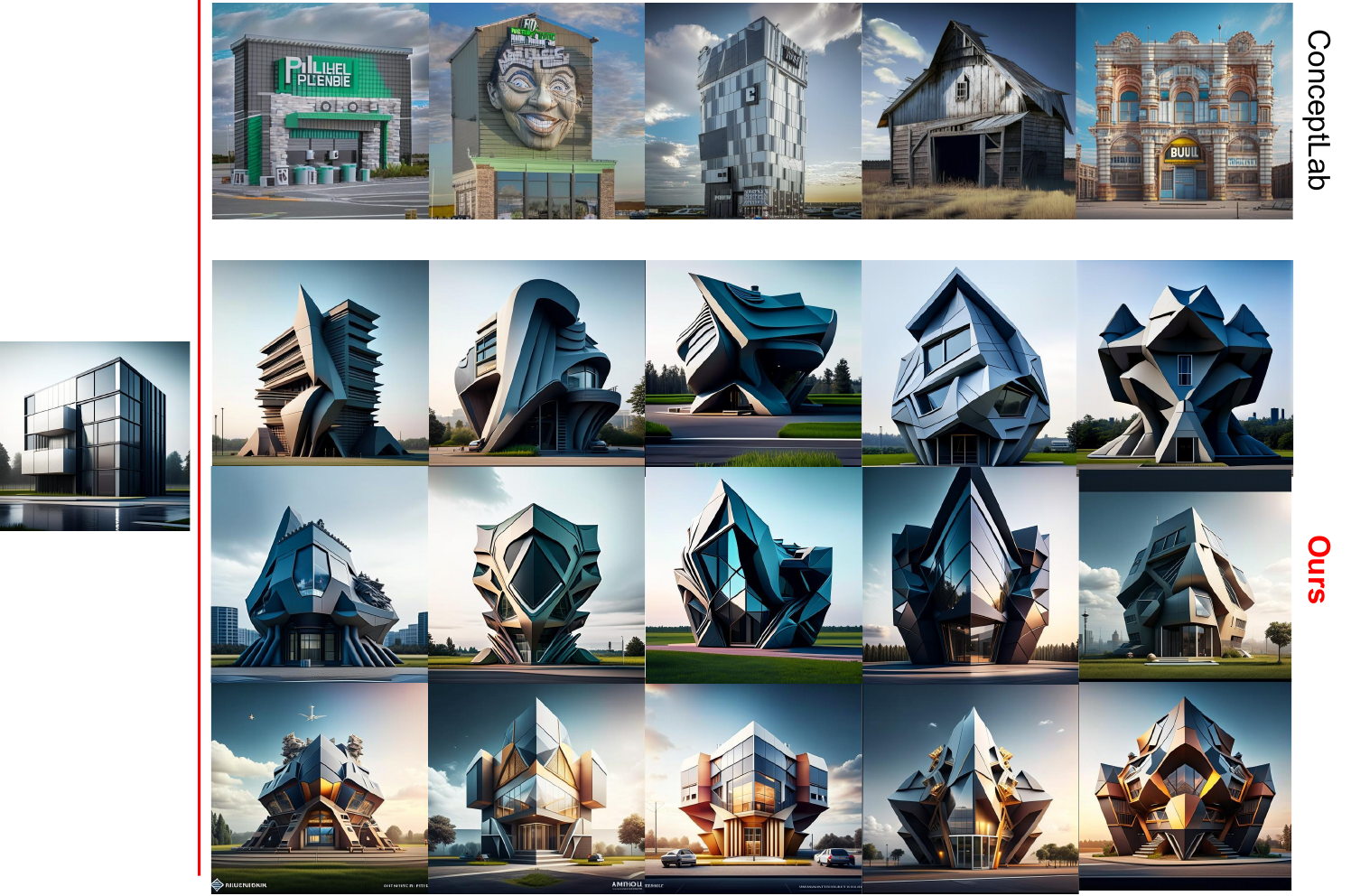}
\caption{Compare with baseline on buildings. Random samples.}
\label{supp_100}
\end{center}
\end{figure*}

\begin{figure*}[ht]
\begin{center}
\includegraphics[width=0.85\linewidth]{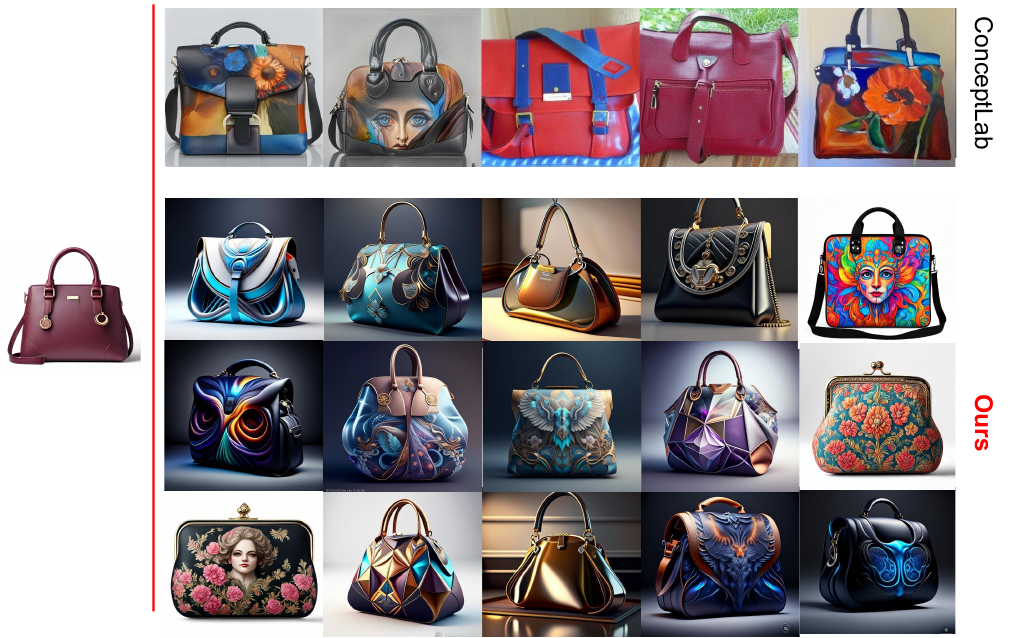}
\caption{Compare with baseline on handbags. Random samples.}
\label{supp_100}
\end{center}
\end{figure*}

\begin{figure*}[ht]
\begin{center}
\includegraphics[width=0.85\linewidth]{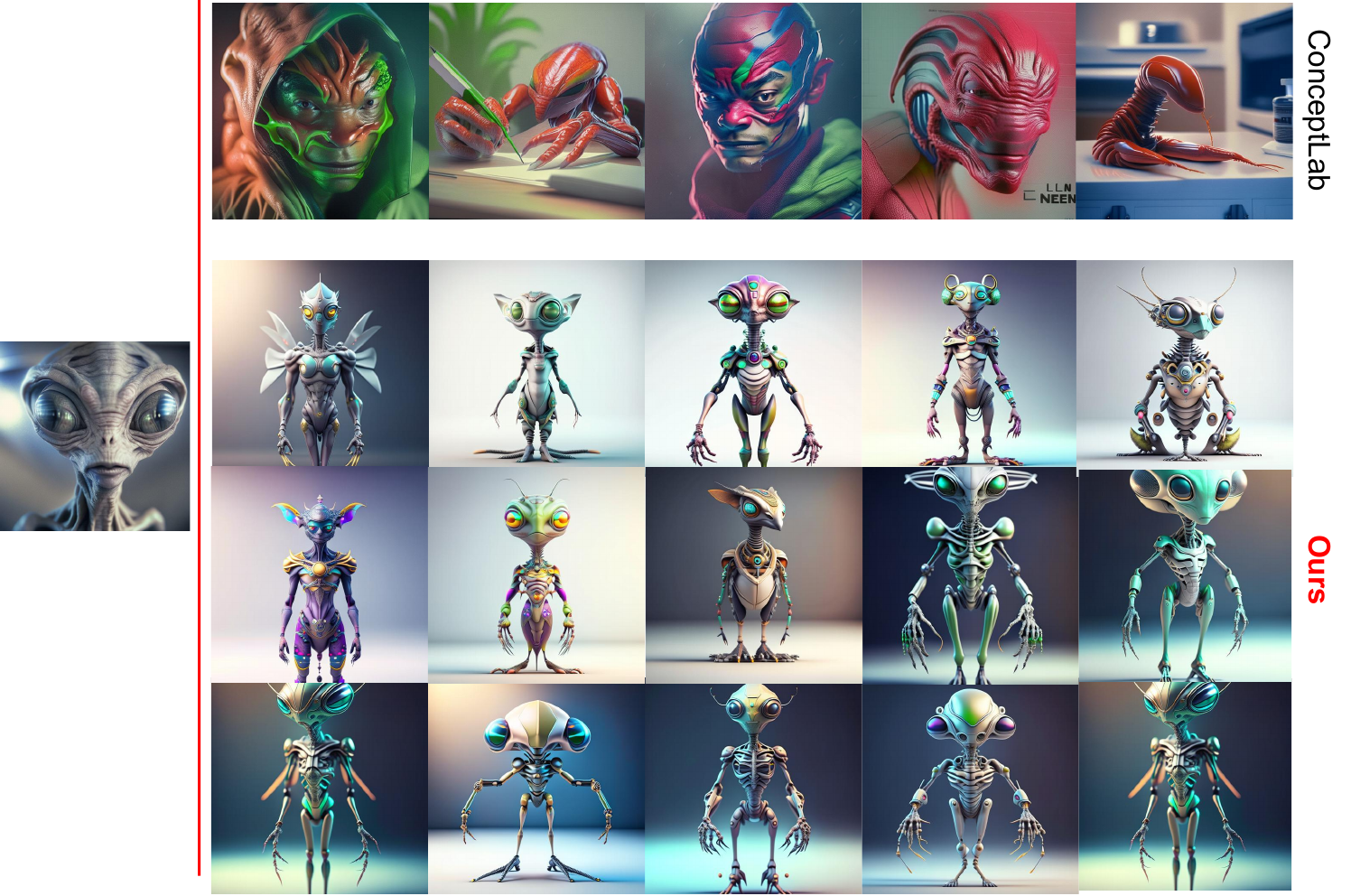}
\caption{Compare with baseline on aliens. Random samples.}
\label{supp_101}
\end{center}
\end{figure*}

\begin{figure*}[ht]
\begin{center}
\includegraphics[width=0.85\linewidth]{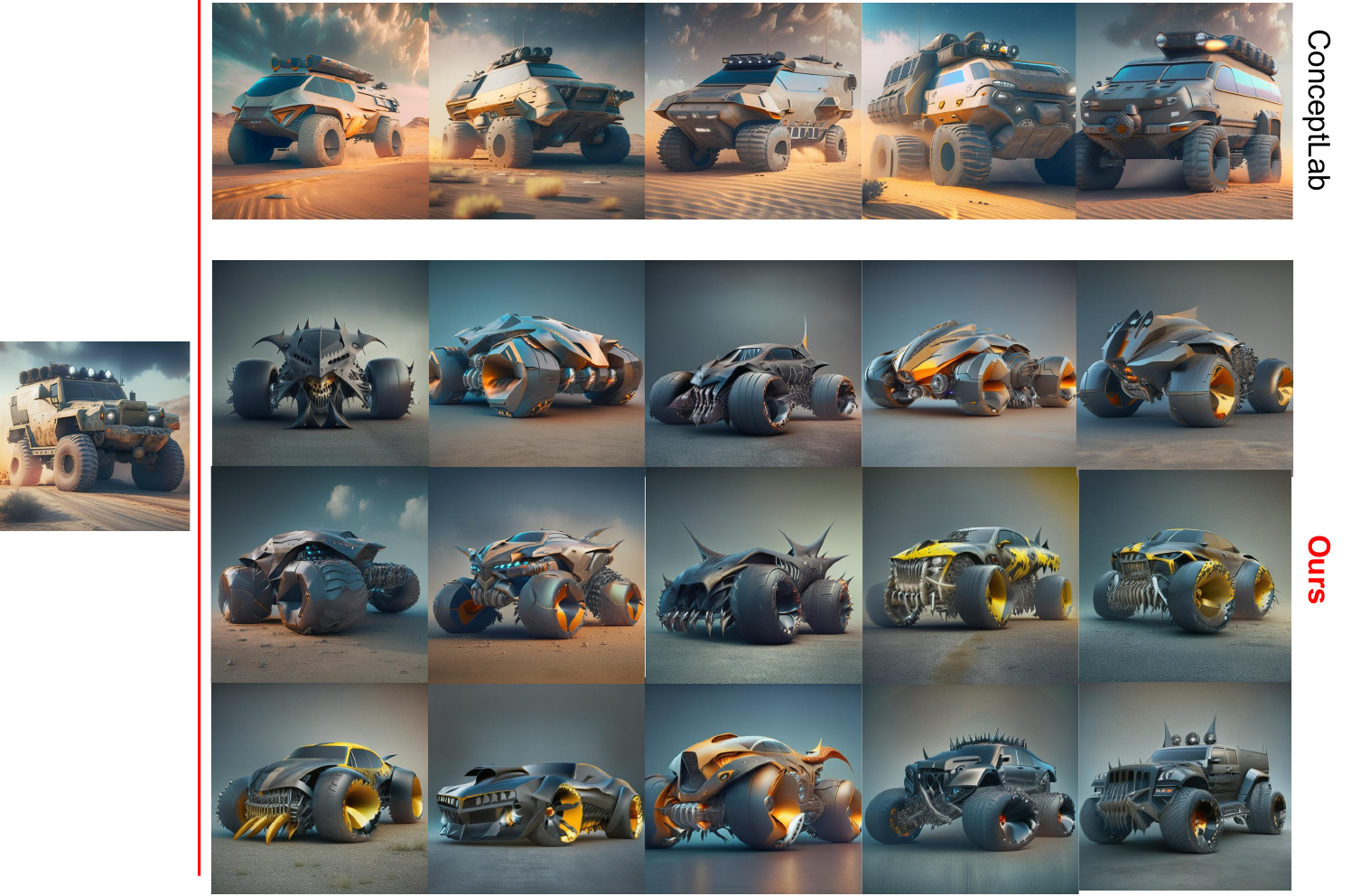}
\caption{Compare with baseline on vehicle. Random samples.}
\label{supp_102}
\end{center}
\end{figure*}

\begin{figure*}[ht]
\begin{center}
\includegraphics[width=0.85\linewidth]{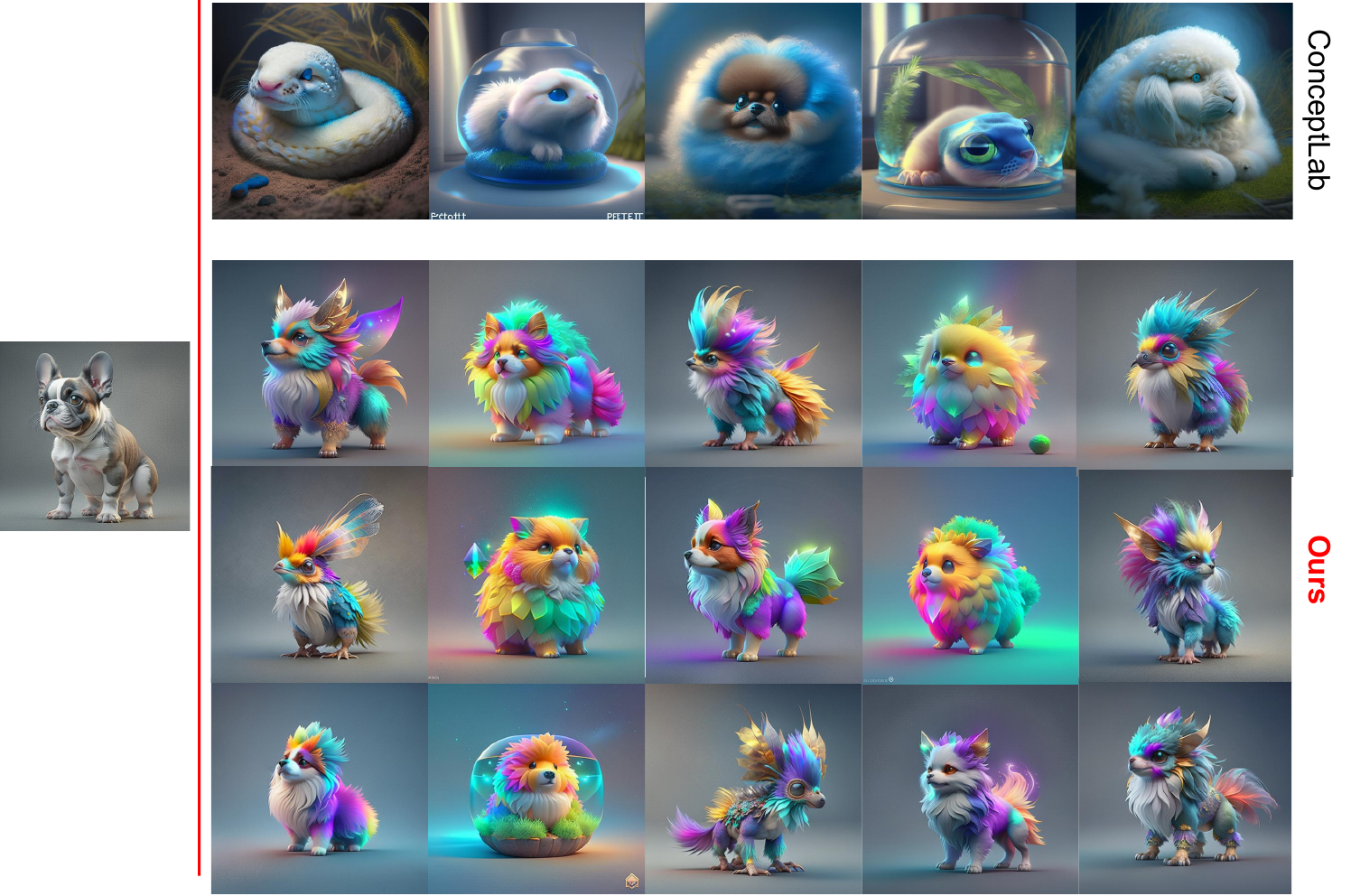}
\caption{Compare with baseline on pet. Random samples.}
\label{supp_103}
\end{center}
\end{figure*}

\begin{figure*}[ht]
\begin{center}
\includegraphics[width=0.85\linewidth]{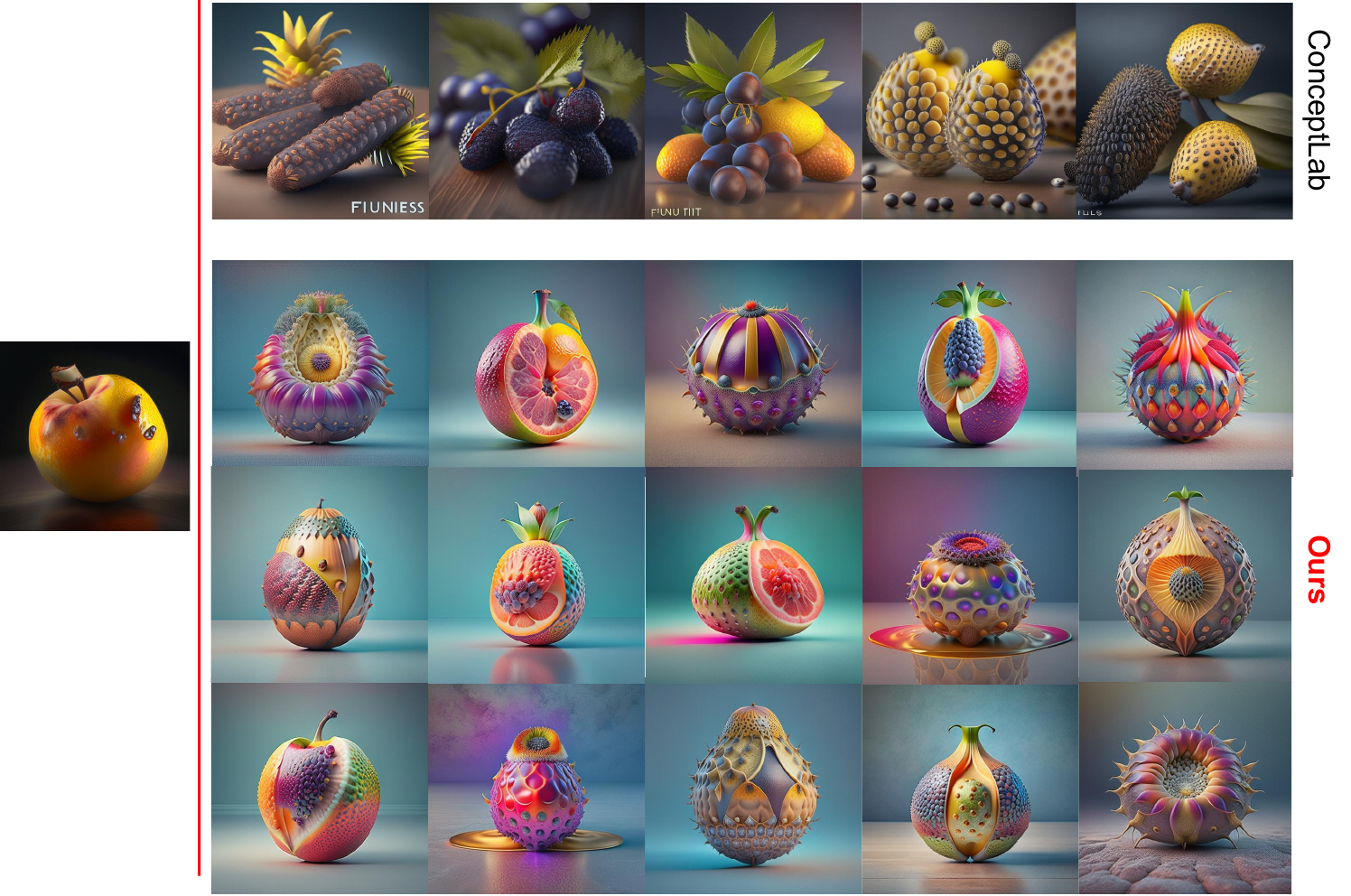}
\caption{Compare with baseline on fruit. Random samples.}
\label{supp_104}
\end{center}
\end{figure*}

\begin{figure*}[ht]
\begin{center}
\includegraphics[width=1\linewidth]{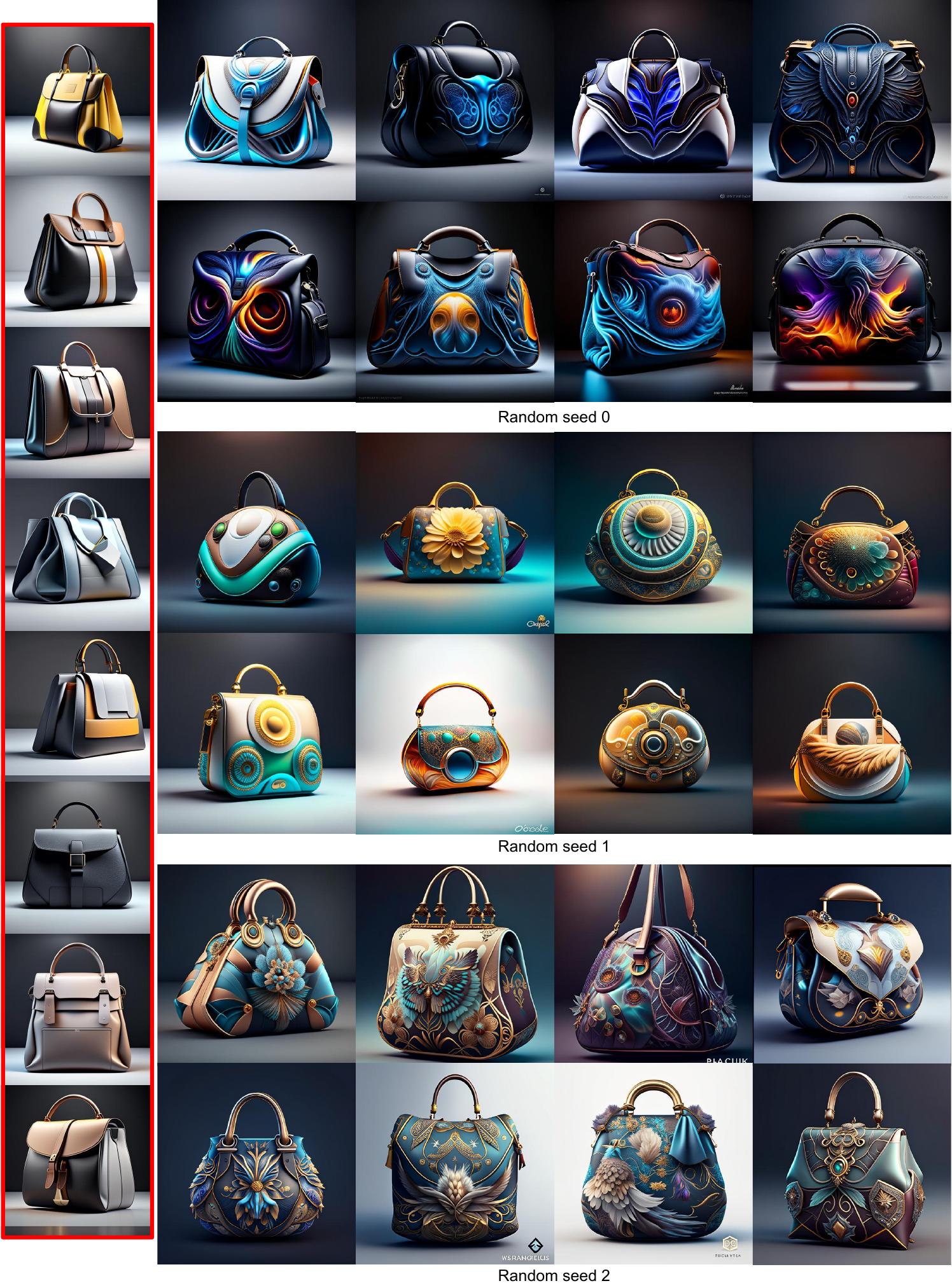}
\caption{creative appendix handbag. (LoRA)}
\label{creative_appendix_handbag}
\end{center}
\end{figure*}

\begin{figure*}[ht]
\begin{center}
\includegraphics[width=1\linewidth]{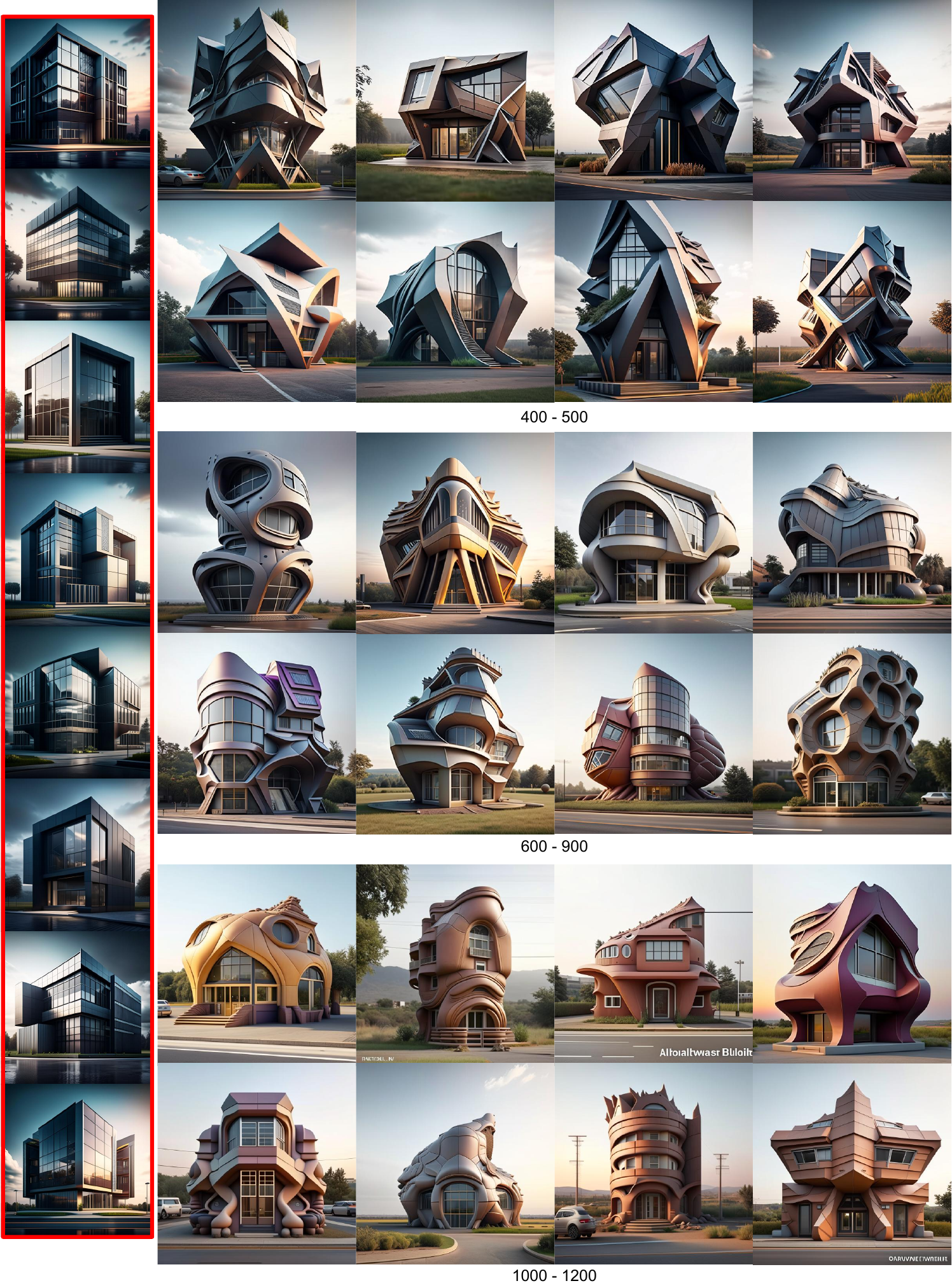}
\caption{creative appendix building. (LoRA)}
\label{creative_appendix_building}
\end{center}
\end{figure*}

\begin{figure*}[ht]
\begin{center}
\includegraphics[width=1\linewidth]{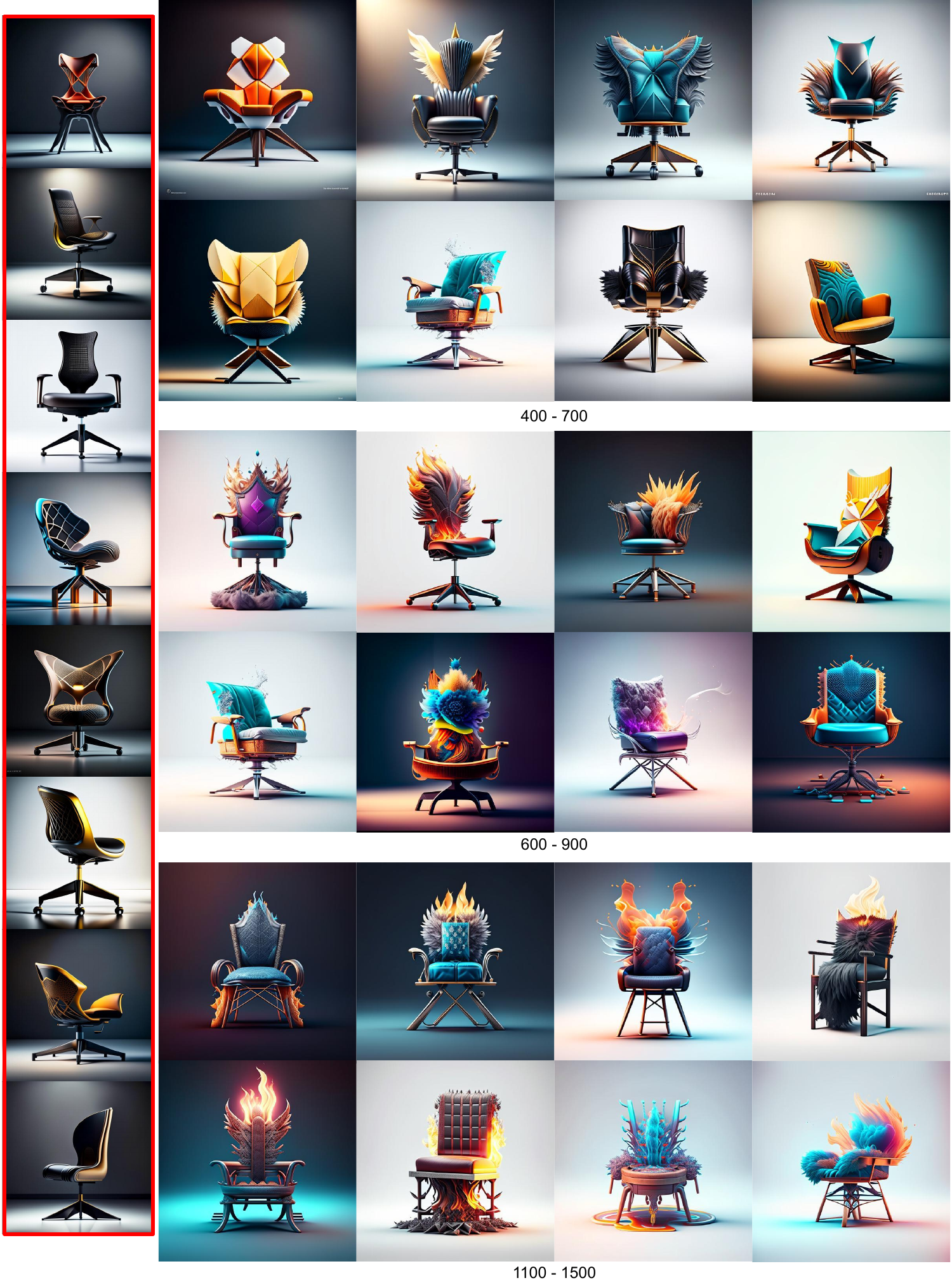}
\caption{creative appendix chair. (LoRA)}
\label{creative_appendix_chair}
\end{center}
\end{figure*}

\begin{figure*}[ht]
\begin{center}
\includegraphics[width=1\linewidth]{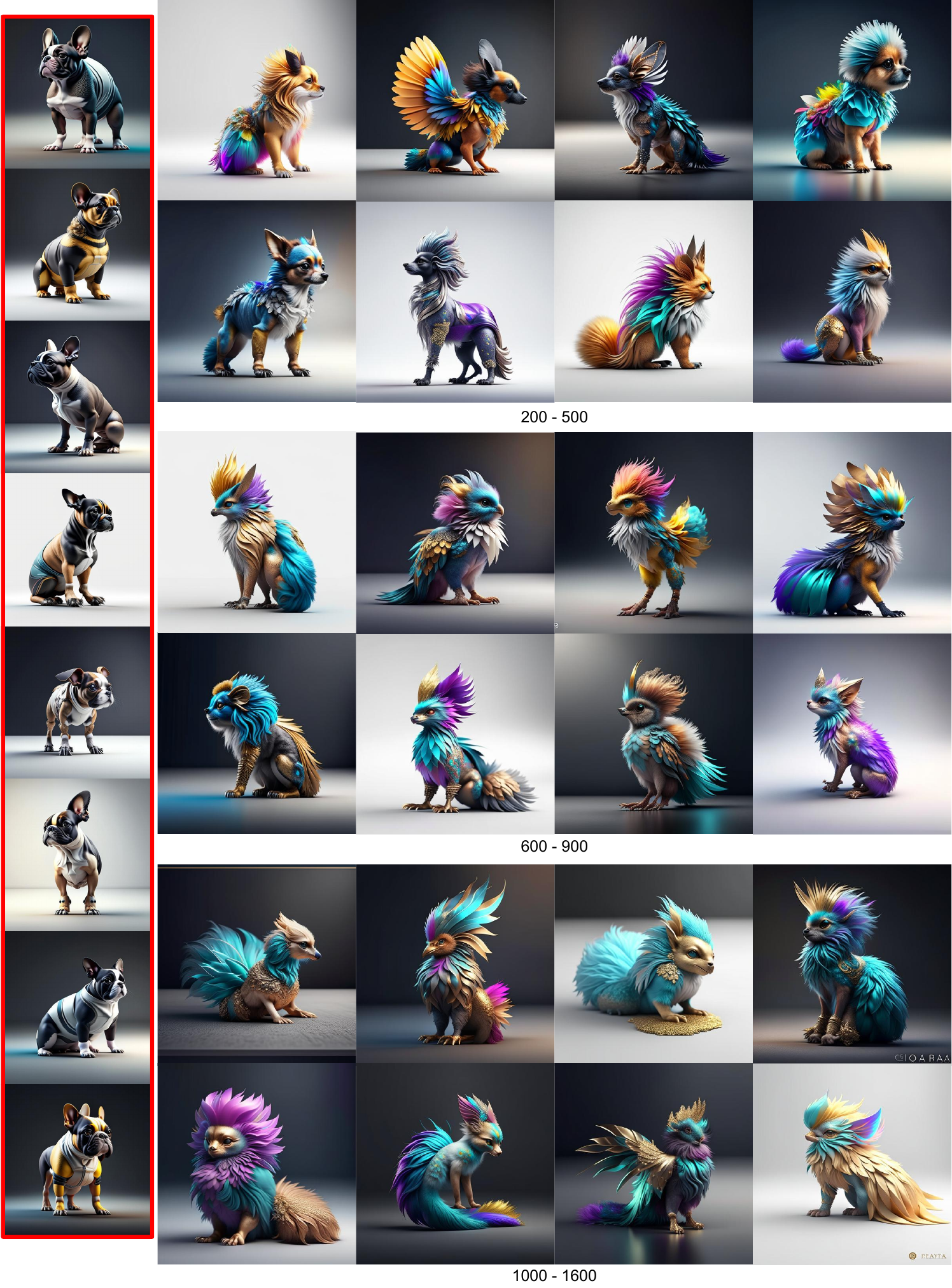}
\caption{creative appendix pet. (LoRA)}
\label{creative_appendix_pet}
\end{center}
\end{figure*}

\begin{figure*}[ht]
\begin{center}
\includegraphics[width=1\linewidth]{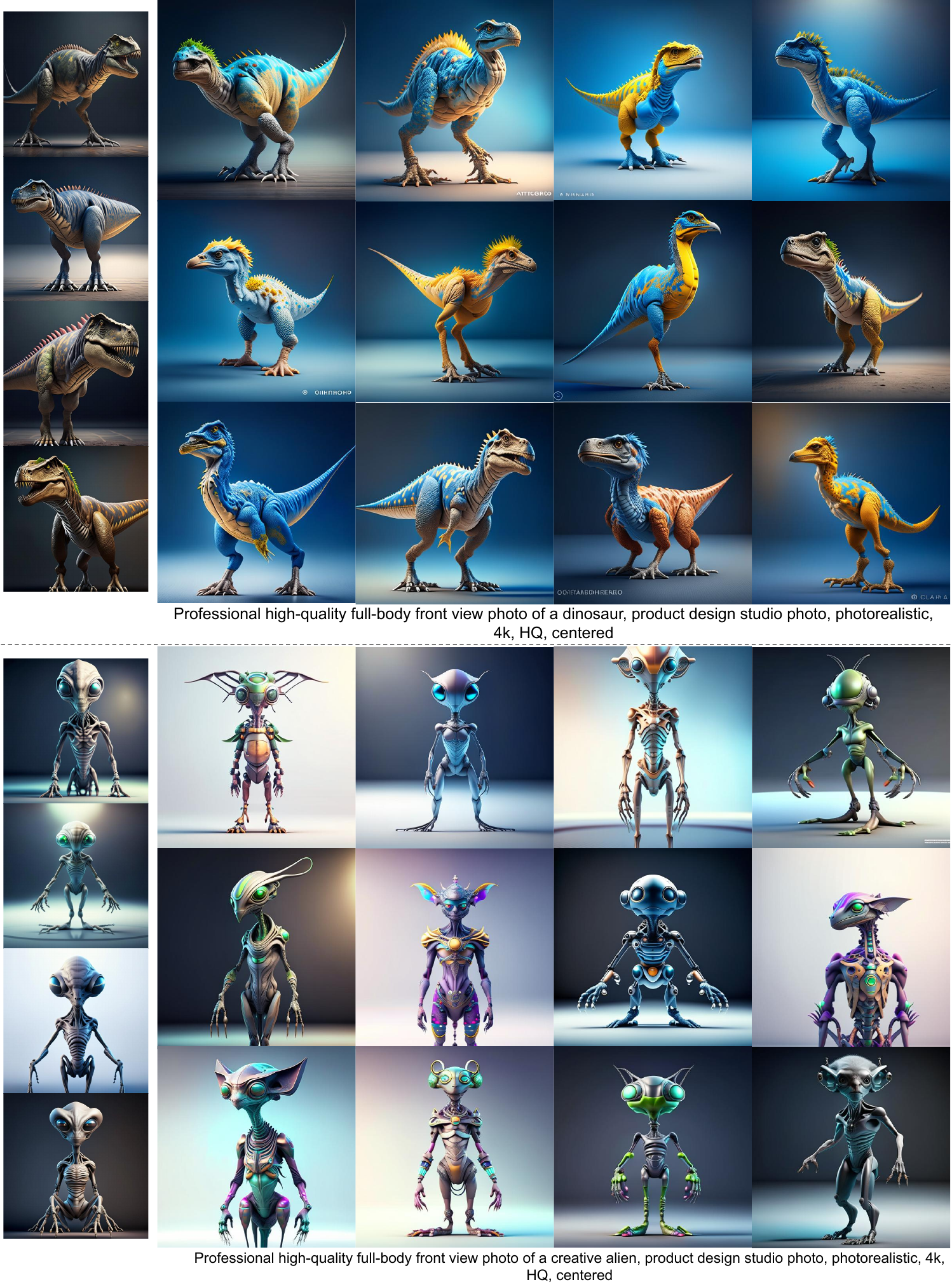}
\caption{creative appendix dinosaur and alien. (LoRA)}
\label{creative_appendix_dinosaurAlien}
\end{center}
\end{figure*}

\clearpage
% \bigskip

% \begin{figure*}[ht]
% \vskip 0.2in
% \begin{center}
% \includegraphics[width=1.0\linewidth]{figures/users.pdf}
% \caption{Sample of our survey.}
% \label{humaneval_1}
% \end{center}
% \vskip -0.2in
% \end{figure*}

\subsection{Human Evaluation. }
\begin{figure}[t]
\centering
\includegraphics[width=\linewidth]{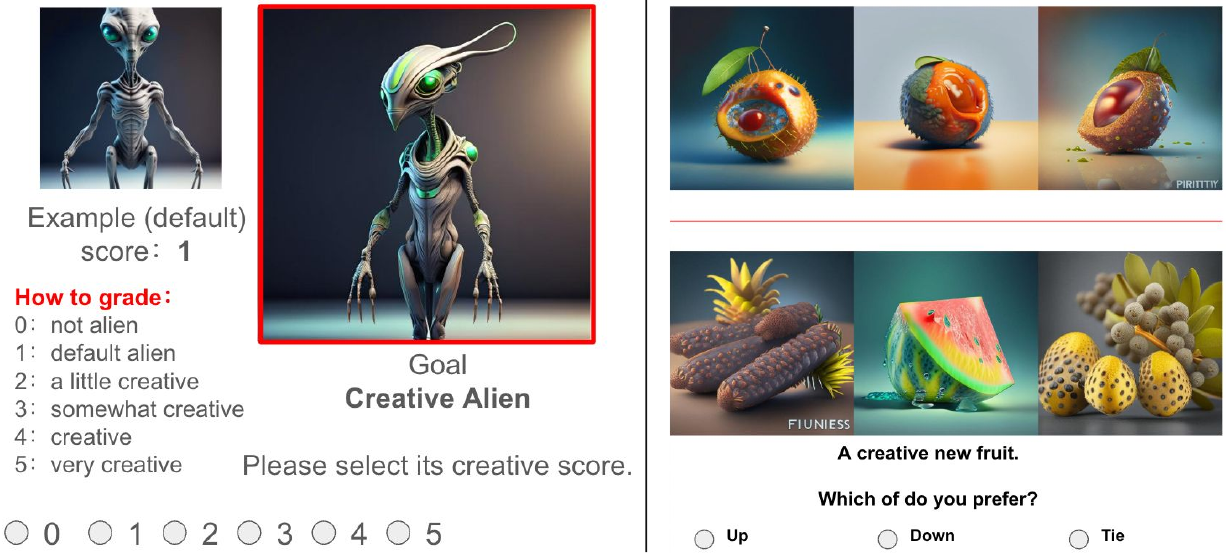}
\caption{Sample of our survey.}
\label{humaneval_1}
\end{figure}

We conducted an extensive human evaluation survey on both Amazon Mechanical Turk and the WJX survey platform, collecting feedback from over 600 users and amassing more than 6000 individual ratings. This large-scale evaluation provided robust quantitative insights into the perceived creativity and semantic fidelity of our generated outputs. A sample of our survey is shown in Fig. \ref{humaneval_1}.

%% file: main.bib
@String(ICLR = {Int. Conf. Learn. Represent.})

@String(AAAI = {AAAI})

@String(ICLR  = {ICLR})

@article{elgammal2017can,
  title={Can: Creative adversarial networks, generating “art” by learning about styles and deviating from style norms},
  author={Elgammal, Ahmed},
  journal={arXiv preprint arXiv:1706.07068},
  volume={6},
  pages={2017},
  year={2017}
}

@article{dhariwal2021diffusion,
  title={Diffusion models beat gans on image synthesis},
  author={Dhariwal, Prafulla and Nichol, Alexander},
  journal={Advances in neural information processing systems},
  volume={34},
  pages={8780--8794},
  year={2021}
}

@article{ho2020denoising,
  title={Denoising diffusion probabilistic models},
  author={Ho, Jonathan and Jain, Ajay and Abbeel, Pieter},
  journal={Advances in neural information processing systems},
  volume={33},
  pages={6840--6851},
  year={2020}
}

@inproceedings{nichol2021improved,
  title={Improved denoising diffusion probabilistic models},
  author={Nichol, Alexander Quinn and Dhariwal, Prafulla},
  booktitle={International conference on machine learning},
  pages={8162--8171},
  year={2021},
  organization={PMLR}
}

@article{nichol2021glide,
  title={Glide: Towards photorealistic image generation and editing with text-guided diffusion models},
  author={Nichol, Alex and Dhariwal, Prafulla and Ramesh, Aditya and Shyam, Pranav and Mishkin, Pamela and McGrew, Bob and Sutskever, Ilya and Chen, Mark},
  journal={arXiv preprint arXiv:2112.10741},
  year={2021}
}

@article{ramesh2022hierarchical,
  title={Hierarchical text-conditional image generation with clip latents},
  author={Ramesh, Aditya and Dhariwal, Prafulla and Nichol, Alex and Chu, Casey and Chen, Mark},
  journal={arXiv preprint arXiv:2204.06125},
  volume={1},
  number={2},
  pages={3},
  year={2022}
}

@inproceedings{rombach2022high,
  title={High-resolution image synthesis with latent diffusion models},
  author={Rombach, Robin and Blattmann, Andreas and Lorenz, Dominik and Esser, Patrick and Ommer, Bj{\"o}rn},
  booktitle={Proceedings of the IEEE/CVF conference on computer vision and pattern recognition},
  pages={10684--10695},
  year={2022}
}

@article{saharia2022photorealistic,
  title={Photorealistic text-to-image diffusion models with deep language understanding},
  author={Saharia, Chitwan and Chan, William and Saxena, Saurabh and Li, Lala and Whang, Jay and Denton, Emily L and Ghasemipour, Kamyar and Gontijo Lopes, Raphael and Karagol Ayan, Burcu and Salimans, Tim and others},
  journal={Advances in neural information processing systems},
  volume={35},
  pages={36479--36494},
  year={2022}
}

@misc{kandinsky2,
  author       = {Sber AI},
  title        = {Kandinsky-2: AI-Driven Image Generation Model},
  year         = {2023},
  howpublished = {\url{https://github.com/ai-forever/Kandinsky-2}},
  note         = {Accessed: 2025-01-24}
}

@misc{kandinsky3,
  author       = {Sber AI},
  title        = {Kandinsky-3: Text-to-image diffusion model},
  year         = {2024},
  howpublished = {\url{https://github.com/ai-forever/Kandinsky-3}},
  note         = {Accessed: 2025-01-24}
}

@article{couairon2022diffedit,
  title={Diffedit: Diffusion-based semantic image editing with mask guidance},
  author={Couairon, Guillaume and Verbeek, Jakob and Schwenk, Holger and Cord, Matthieu},
  journal={arXiv preprint arXiv:2210.11427},
  year={2022}
}

@article{hertz2022prompt,
  title={Prompt-to-prompt image editing with cross attention control},
  author={Hertz, Amir and Mokady, Ron and Tenenbaum, Jay and Aberman, Kfir and Pritch, Yael and Cohen-Or, Daniel},
  journal={arXiv preprint arXiv:2208.01626},
  year={2022}
}

@article{meng2021sdedit,
  title={Sdedit: Guided image synthesis and editing with stochastic differential equations},
  author={Meng, Chenlin and He, Yutong and Song, Yang and Song, Jiaming and Wu, Jiajun and Zhu, Jun-Yan and Ermon, Stefano},
  journal={arXiv preprint arXiv:2108.01073},
  year={2021}
}

@article{podell2023sdxl,
  title={Sdxl: Improving latent diffusion models for high-resolution image synthesis},
  author={Podell, Dustin and English, Zion and Lacey, Kyle and Blattmann, Andreas and Dockhorn, Tim and M{\"u}ller, Jonas and Penna, Joe and Rombach, Robin},
  journal={arXiv preprint arXiv:2307.01952},
  year={2023}
}

@inproceedings{ramesh2021zero,
  title={Zero-shot text-to-image generation},
  author={Ramesh, Aditya and Pavlov, Mikhail and Goh, Gabriel and Gray, Scott and Voss, Chelsea and Radford, Alec and Chen, Mark and Sutskever, Ilya},
  booktitle={International conference on machine learning},
  pages={8821--8831},
  year={2021},
  organization={Pmlr}
}

@article{richardson2023conceptlab,
  title={Conceptlab: Creative generation using diffusion prior constraints},
  author={Richardson, Elad and Goldberg, Kfir and Alaluf, Yuval and Cohen-Or, Daniel},
  journal={arXiv preprint arXiv:2308.02669},
  year={2023}
}

@article{hu2022lora,
  title={Lora: Low-rank adaptation of large language models.},
  author={Hu, Edward J and Shen, Yelong and Wallis, Phillip and Allen-Zhu, Zeyuan and Li, Yuanzhi and Wang, Shean and Wang, Lu and Chen, Weizhu and others},
  journal={ICLR},
  volume={1},
  number={2},
  pages={3},
  year={2022}
}

@article{doi:10.1080/14786440109462720,
author = { Karl   Pearson   F.R.S. },
title = {LIII. On lines and planes of closest fit to systems of points in space},
journal = {The London, Edinburgh, and Dublin Philosophical Magazine and Journal of Science},
volume = {2},
number = {11},
pages = {559-572},
year  = {1901},
publisher = {Taylor & Francis},
doi = {10.1080/14786440109462720},
}

@article{Goodfellow2014,
  author = {Goodfellow, I. and Pouget-Abadie, J. and Mirza, M. and Xu, B. and Warde-Farley, D. and Ozair, S. and Bengio, Y.},
  title = {Generative adversarial nets},
  journal = {Advances in Neural Information Processing Systems},
  volume = {27},
  year = {2014}
}

@article{Brock2018,
  author = {Brock, A. and Donahue, J. and Simonyan, K.},
  title = {Large scale GAN training for high fidelity natural image synthesis},
  journal = {arXiv preprint arXiv:1809.11096},
  year = {2018}
}

@article{Kingma2013,
  author = {Kingma, D. P. and Welling, M.},
  title = {Auto-encoding variational Bayes},
  journal = {arXiv preprint arXiv:1312.6114},
  year = {2013}
}

@inproceedings{kingma2018glow,
  author    = {Kingma, Diederik P. and Dhariwal, Prafulla},
  title     = {Glow: Generative Flow with Invertible 1x1 Convolutions},
  booktitle = {Advances in Neural Information Processing Systems},
  volume    = {31},
  year      = {2018}
}

@article{dinh2014nice,
  author    = {Dinh, Laurent and Krueger, David and Bengio, Yoshua},
  title     = {NICE: Non-linear Independent Components Estimation},
  journal   = {arXiv preprint},
  volume    = {arXiv:1410.8516},
  year      = {2014},
  eprint    = {1410.8516},
  archivePrefix = {arXiv}
}

@article{song2020score,
  author    = {Song, Yang and Sohl-Dickstein, Jascha and Kingma, Diederik P. and Kumar, Abhishek and Ermon, Stefano and Poole, Ben},
  title     = {Score-Based Generative Modeling through Stochastic Differential Equations},
  journal   = {arXiv preprint},
  volume    = {arXiv:2011.13456},
  year      = {2020},
  eprint    = {2011.13456},
  archivePrefix = {arXiv}
}

@article{ho2022classifier,
  author    = {Ho, Jonathan and Salimans, Tim},
  title     = {Classifier-Free Diffusion Guidance},
  journal   = {arXiv preprint},
  volume    = {arXiv:2207.12598},
  year      = {2022},
  eprint    = {2207.12598},
  archivePrefix = {arXiv}
}

@inproceedings{heusel2017gans,
  author    = {Heusel, Martin and Ramsauer, Hubert and Unterthiner, Thomas and Nessler, Bernhard and Hochreiter, Sepp},
  title     = {GANs Trained by a Two Time-Scale Update Rule Converge to a Local Nash Equilibrium},
  booktitle = {Advances in Neural Information Processing Systems},
  volume    = {30},
  year      = {2017}
}

@article{theis2015note,
  author    = {Theis, Lucas and van den Oord, Aäron and Bethge, Matthias},
  title     = {A Note on the Evaluation of Generative Models},
  journal   = {arXiv preprint},
  volume    = {arXiv:1511.01844},
  year      = {2015},
  eprint    = {1511.01844},
  archivePrefix = {arXiv}
}

@inproceedings{salimans2016improved,
  author    = {Salimans, Tim and Goodfellow, Ian and Zaremba, Wojciech and Cheung, Vicki and Radford, Alec and Chen, Xi},
  title     = {Improved Techniques for Training GANs},
  booktitle = {Advances in Neural Information Processing Systems},
  volume    = {29},
  year      = {2016}
}

@incollection{machado2008iterative,
  author    = {Machado, Penousal and Romero, Juan and Manaris, Bilal},
  title     = {An Iterative Approach to Stylistic Change in Evolutionary Art},
  booktitle = {The Art of Artificial Evolution: A Handbook on Evolutionary Art and Music},
  publisher = {Springer},    
  year      = {2008}
}

@inproceedings{dipaola2007incorporating,
  author    = {DiPaola, Steve R. and Gabora, Liane},
  title     = {Incorporating Characteristics of Human Creativity into an Evolutionary Art Algorithm},
  booktitle = {Proceedings of the 9th Annual Conference Companion on Genetic and Evolutionary Computation},
  year      = {2007},
  pages     = {2450--2456},
  month     = {July}
}

@inproceedings{baker1994evolving,
  author    = {Baker, Eric and Seltzer, Margo},
  title     = {Evolving Line Drawings},
  booktitle = {Proceedings of Graphics Interface '94},
  year      = {1994},
  location  = {Banff, Canada},
  pages     = {91--100},
  month     = {May}
}

@incollection{graf1995interactive,
  author    = {Graf, J{\"o}rg and Banzhaf, Wolfgang},
  title     = {Interactive Evolution of Images},
  booktitle = {Evolutionary Programming},
  pages     = {53--65},
  year      = {1995},
  month     = {March}
}

@inproceedings{colton2008creativity,
  author    = {Colton, Simon},
  title     = {Creativity Versus the Perception of Creativity in Computational Systems},
  booktitle = {AAAI Spring Symposium: Creative Intelligent Systems},
  volume    = {8},
  pages     = {7},
  year      = {2008},
  month     = {March}
}

@inproceedings{Berlyne1967,
  author = {Berlyne, D. E.},
  title = {Arousal and reinforcement},
  booktitle = {Nebraska symposium on motivation},
  publisher = {University of Nebraska Press},
  year = {1967}
}

@book{Berlyne1971,
  author = {Berlyne, D. E.},
  title = {Aesthetics and Psychobiology},
  publisher = {Appleton-Century-Crofts},
  address = {New York},
  year = {1971}
}

@book{Boden2004,
  author = {Boden, M. A.},
  title = {The Creative Mind: Myths and Mechanisms},
  publisher = {Psychology Press},
  year = {2004}
}

@book{Boden2010,
  author = {Boden, M. A.},
  title = {Creativity and Art, Three Roads to Surprise},
  publisher = {Oxford University Press},
  year = {2010}
}

@article{schneirla1959evolutionary,
  author    = {Schneirla, T. C.},
  title     = {An Evolutionary and Developmental Theory of Biphasic Processes Underlying Approach and Withdrawal},
  journal   = {Psychological Review},
  year      = {1959}
}

@book{wundt1874grundzuge,
  author    = {Wundt, Wilhelm M.},
  title     = {Grundzüge der physiologischen Psychologie},
  publisher = {W. Engelmann},
  year      = {1874}
}

@article{kuppens2013relation,
  author    = {Kuppens, Peter and Tuerlinckx, Francis and Russell, James A. and Barrett, Lisa Feldman},
  title     = {The Relation Between Valence and Arousal in Subjective Experience},
  journal   = {Psychological Bulletin},
  volume    = {139},
  number    = {4},
  pages     = {917--940},
  year      = {2013},
  month     = {July},
  doi       = {10.1037/a0030811}
}

@inproceedings{esser2024scaling,
  title={Scaling rectified flow transformers for high-resolution image synthesis},
  author={Esser, Patrick and Kulal, Sumith and Blattmann, Andreas and Entezari, Rahim and M{\"u}ller, Jonas and Saini, Harry and Levi, Yam and Lorenz, Dominik and Sauer, Axel and Boesel, Frederic and others},
  booktitle={Forty-first international conference on machine learning},
  year={2024}
}

@article{xiong2024novel,
  title={Novel Object Synthesis via Adaptive Text-Image Harmony},
  author={Xiong, Zeren and Zhang, Zedong and Chen, Zikun and Chen, Shuo and Li, Xiang and Sun, Gan and Yang, Jian and Li, Jun},
  journal={arXiv preprint arXiv:2410.20823},
  year={2024}
}
